\documentclass[journal]{IEEEtran}

\ifCLASSINFOpdf
\else
   \usepackage[dvips]{graphicx}
\fi
\usepackage{url}
\usepackage{blindtext}
\usepackage{graphicx}
\usepackage{amsmath,amssymb}
\usepackage{lineno}
\usepackage{color}
\usepackage{subfigure}
\usepackage{multirow}
\usepackage{algorithm}
\usepackage{fancyhdr}
\usepackage{cases}
\usepackage{epsfig}
\usepackage{epstopdf}
\usepackage{array,color}
\usepackage{amsmath}
\usepackage{stfloats}
\usepackage{url}
\usepackage{algorithm}
\usepackage{algorithmic}
\usepackage{amsfonts,amssymb}
\usepackage{amsmath}
\usepackage{bm}
\usepackage{booktabs}
\usepackage{array}
\usepackage{mathrsfs}
\usepackage{graphicx}
\usepackage{cite}
\usepackage{stfloats}
\usepackage{float}
\usepackage{eqnarray}
\usepackage{hyperref}
\usepackage{makecell}
\hypersetup{hidelinks}
\usepackage{booktabs}

\begin{document}

\title{Depth-induced Saliency Comparison Network for Diagnosis of Alzheimer's Disease via Jointly Analysis of Visual Stimuli and Eye Movements}


\author{Yu~Liu,~\IEEEmembership{Member,~IEEE,} 
        Wenlin~Zhang,
        Shaochu~Wang,
        Fangyu~Zuo,
        Peiguang~Jing,
        Yong~Ji
\thanks{Yu~Liu is with the School of Microelectronics and Zhejiang International Institute of Innovative Design and International Institute for Innovative Design and Intelligent Manufacturing of Tianjin University in Zhejiang, China. 
Wenlin~Zhang received the B.S. and M.S. degrees in communication engineering from Tianjin University, Tianjin, China, in 2020 and 2023, respectively.  Where he is currently pursuing the Ph.D. degree in circuits and systems. 
Shaochu~Wang received the B.E. degree in electronics engineering from China University of Geosciences, Beijing, China, in 2007 and the M.S. degree in signal and information processing and the Ph.D. degree in circuits and systems from Tianjin University, Tianjin, China, in 2009 and 2014, respectively.He was a Visiting Scholar with the Department of Computer Science and Engineering, Lehigh University, Bethlehem, PA, USA, from 2011 to 2012.
Fangyu~Zuo is with the School of Microelectronics, Tianjin University, Tianjin, China.
Peiguang Jing is with School of Electrical and Information Engineering, Tianjin University, Tianjin, China.
Yong~Ji is with Tianjin Key Laboratory of Cerebrovascular and Neurodegenerative Diseases, Department of Neurology, Tianjin Dementia Institute, Tianjin Huanhu Hospital, Tianjin, China.

\textit{Corresponding author: Yu Liu, liuyu@tju.edu.cn.}} }

\markboth{Journal of ***}
{Shell \MakeLowercase{\textit{et al.}}: Bare Demo of IEEEtran.cls for IEEE Journals}
\maketitle

\begin{abstract}
Early diagnosis of Alzheimer's Disease (AD) is very important for following medical treatments, and eye movements under special visual stimuli may serve as a potential non-invasive biomarker for detecting cognitive abnormalities of AD patients. In this paper, we propose an Depth-induced saliency comparison network (DISCN) for eye movement analysis, which may be used for diagnosis the Alzheimers disease. In DISCN, a salient attention module fuses normal eye movements with RGB and depth maps of visual stimuli using hierarchical salient attention (SAA) to evaluate comprehensive saliency maps, which contain information from both visual stimuli and normal eye movement behaviors. In addition, we introduce serial attention module (SEA) to emphasis the most abnormal eye movement behaviors to reduce personal bias for a more robust result. According to our experiments, the DISCN achieves consistent validity in classifying the eye movements between the AD patients and normal controls.  
\end{abstract}

\IEEEpeerreviewmaketitle

\section{Introduction}

\IEEEPARstart{A}{lzheimer's} disease (AD) is the leading cause of dementia, which gradually impairs cognitive abilities such as memory, language, and mind of the elderly. Timely diagnosis of AD is critical, as it offers the opportunity for early intervention and more aggressive treatments \cite{dubois2016timely}. Traditional diagnostic methods, including cognitive evaluations \cite{chen2017early}, daily activity monitoring \cite{palacios2022cognitive}, blood tests \cite{eke2020early}, and medical image analysis \cite{zhou2014multivariate}, have limited efficiency due to their reliance on experienced medical experts and complexity. For example, cognitive evaluations like Mini‐Mental State Examination (MMSE) \cite{liu2020enhancing} requires experienced medical experts to spend 10-15 minutes on each individual for consultation, while obtaining magnetic resonance imaging (MRI) and blood samples may cause invasive damage to patients. Thus, researchers are seeking new biomarkers to achieve more friendly and efficient diagnostic approaches in the early stage of AD.

Recent studies indicate that eye movements is a promising biomarker for diagnosing early AD, which is sensitive to cognitive decline caused by AD \cite{noiret2017saccadic}. Consequently, eye movements have the potential to be an indicator of AD onset \cite{readman2021potential}\cite{coors2022associations}. For example, Opwonya \textit{et. al} \cite{opwonya2022saccadic} reviewed literature related to AD diagnosis with eye movement features, and concluded that prosaccade antisaccade latencies and frequency of antisaccade errors showed significant potential in for AD diagnosis; Parra \textit{et. al} \cite{parra2022memory} analyzed eye movement behaviors during the visual short-term memory binding task (VSTMB), and suggested that patients performed abnormal saccades and fixation durations indicating impairments in memory and executive functions; Ramzaoui \textit{et. al} \cite{ramzaoui2022top} used scenes depicting real environments to analyse eye-movement differences between AD patients and normal controls, and they concluded that AD patients had longer search times, and showed a greater probability of distractor selection during the trials. All the above studies have shown great promise that eye movements can reflect the cognitive impairments in AD patients.

To improve diagnostic efficiency and achieve earlier AD detection, more and more researchers are using artifitial intelligence (AI) to process various diagnostic data for AD. Among the data, medical images \cite{liu2020enhancing,shi2022unsupervised} and scales \cite{varma2023early} are the most common inputs for deep learning networks. But recent studies have also incorporated eye movements to differentiate AD patients from normal controls. Tsai \textit{et. al} \cite{tsai2021machine} processed the eye tracking and navigation data collected from games based on virtual reality (VR) with support vector machine (SVM), random forest (RF) and tree structures; Przybyszewski \textit{et. al} \cite{przybyszewski2023machine} reviewed literature that used machine learning for the diagnosis of AD through high-dimensional data such as eye movements, neurological and psychological trials. 

With the development of deep learning (DL) and eye tracking equipments, deep learning-based networks for diagnosing AD by eye movements collected from various eye tracking trials are developing. Haque \textit{et. al} \cite{haque2020deep} implemented passive visual memory tests on mobile iPad devices and extracted diagnostic features from eye movements via deep convolution neural network (CNN) and transfer learning; Sun \textit{et. al} \cite{sun2022novel} explored key eye movement features associated with AD and built a deep learning network to classify the eye movements collected during a three-dimensional (3D) visual paired comparison (VPC) task; Vinayak \textit{et. al} \cite{vinayak2023prediction} performed a scan path experiment and carried a machine learning approach to encode multidimensional data consisting scan length, angle and radical maps to obtain classification results for AD diagnosis; Yin \textit{et. al} \cite{yin2023internet} proposed an Internet of Things (IoT) architecture constructed with deep learning that provided automatic identification of early-stage AD and distinguished AD patients from normal controls. However, most of these studies focus on eye-movement features such as saccade, anti-saccade, and visual memory abilities. Exploration of visual saliency features is relatively limited. 


\begin{figure}[t]
\centerline{\includegraphics[width=1.02\columnwidth]{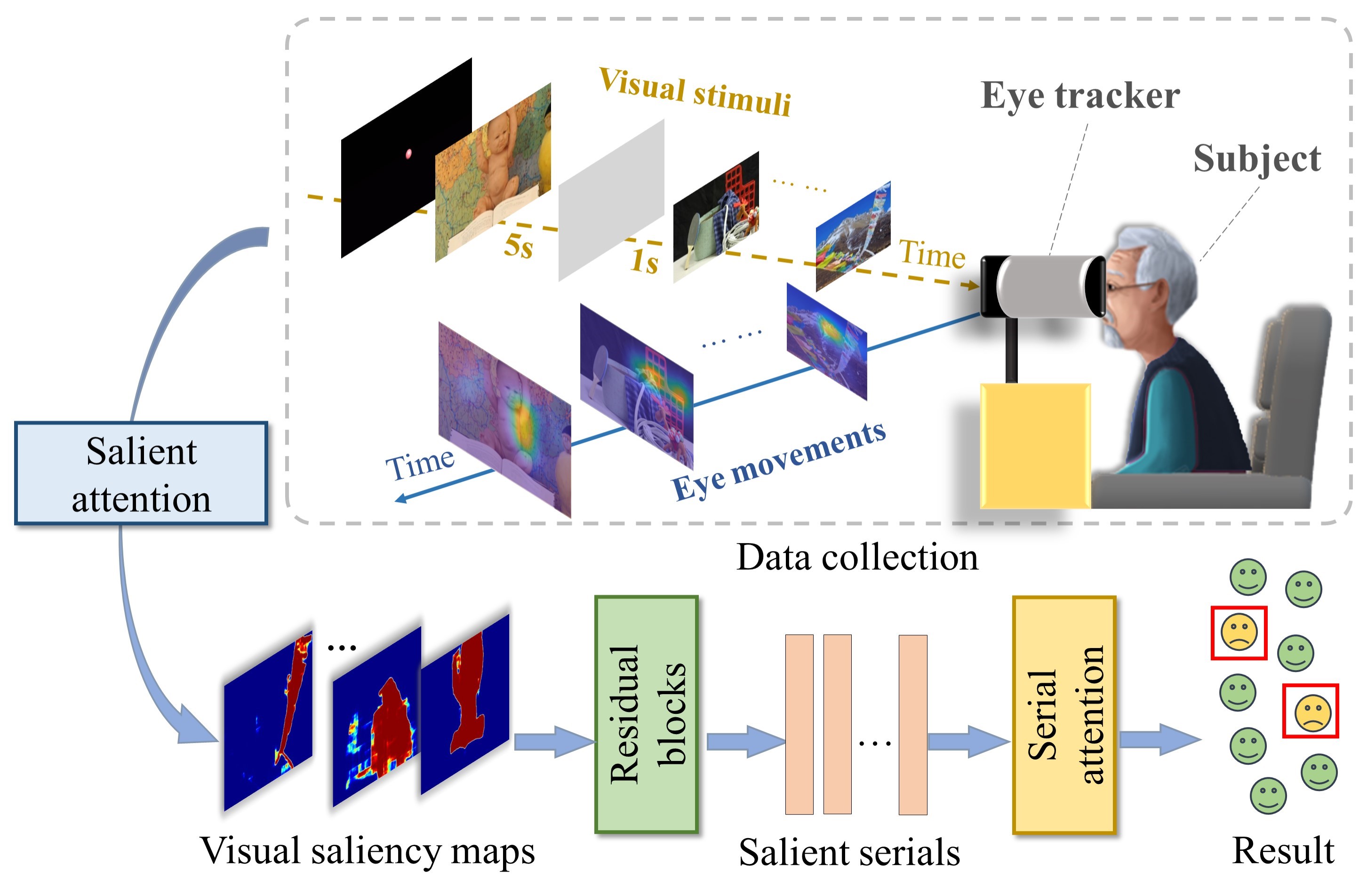}}
\caption{The overall process of the proposed approach for diagnosing AD with visual salience evaluation and deep learning.}
\vspace{-1em}
\label{process of data collection}
\end{figure}

In this paper, we propose a novel approach to diagnose AD using deep learning-based network named Depth-induced Integrated Comparison serial attention Network (DISCN) with eye movements collected during a free viewing task. As shown in Fig. \ref{process of data collection}, a series of images in left and right eye perspectives are presented to subjects binocularly by an 3D eye tracker designed by Sun \textit{et. al} \cite{sun2023}, which records eye movements of subjects and transforms eye movements into heatmaps. Then, 3D visual stimuli and heatmaps are integrated into visual saliecy maps by an integration module. The visual saliency maps are processed by a serial attention module to get a diagnosis result. The major contributions of this paper are summarized as follows:

\begin{itemize} 
    \item Targeting at the diagnosis of AD with eye movement data, we proposed a novel deep learning-based DISCN, which contains an integration module to fuse visual stimuli with eye movements for evaluating visual saliency maps comprehensively. serial attention module is applied to extract visual saliency features from the visual saliency maps, so that the AD detecting accuracy is enhanced.
    \item We conducted extensive experiments on a collected eye movements dataset and completed comprehensive comparison, The results showed our approach outperformed the state-of-the-art models. 
\end{itemize}

\section{Related Work}

\subsection{AD diagnosing methods based on eye movements}


Saccade eye movement including saccade latency, saccade errors \cite{eraslan2023uncorrected, hannonen2022shortening,eraslan2022influence} are commonly used for AD diagnosis. It aims to establish an evaluation criteria for the detection of abnormal eye-tracking behavior in AD patients for the initial diagnosis of AD patients by well-designed clinical eye-tracking tests. Eraslan \textit{et. al} \cite{eraslan2023uncorrected} designed a saccade task that is applied on patients with AD and Mild Cognitive Impairment (MCI), and healthy controls, which found that both AD patients and MCI showed varying degrees of elevated antisaccade error rates and reduced saccade accuracy compared to healthy controls. This suggests that cognitive impairment in AD patients is highly correlated with abnormal saccade performances, which gives a promising eye-movement indicator for the initial diagnosis of AD patients. 

To simplify the process of saccade tasks and provide a more natural experience, Hannonen \textit{et. al} \cite{hannonen2022shortening} designed a shorter reading task that detected subjects' saccade behaviors including saccade speed, saccade amplitude, and saccade frequency during reading, which found that AD patients performed worse in the reading test compared with healthy controls. To further confirm the correlation between the results of the eye hopping task and the traditional medical diagnosis of AD, Eraslan \textit{et. al} \cite{eraslan2022influence} performed saccade tests and neuropsychological tests on AD patients and healthy controls respectively. The result showed that the rate and amplitude of saccade correlated strongly with the neurological test results , which allowed for an accurate diagnosis of AD patients.

In order to get more consistent diagnosis of AD, researchers have explored more diverse eye movement indicators to diagnose AD in addition to saccades, and have designed more comprehensive eye movement tests \cite{tokushige2023early,jang2021classification}. Tokushige \textit{et. al} \cite{tokushige2023early} designed a visual memory task to record and analyze the visual search processes and visual attention of AD patients and healthy controls while performing the task. The results showed that AD patients paid insufficient attention to the informational  parts and required longer time for right visual exploration. This suggested that visual attention and visual exploration processes are also promising eye movement indicators for the diagnosis of AD. In addition, to obtain more comprehensive eye movement diagnostic indicators, Jang \textit{et. al} \cite{jang2021classification} designed a picture depiction task and assessed the subjects from multiple dimensions such as saccade, gaze, visual attention, and verbal ability, which verified that the multimodal eye movement indicators had superior stability for AD diagnosis.

\subsection{Serial attention neural networks}

With the increasing demand for intelligent and efficient AD diagnosis, deep learning methods are increasingly being applied to the analysis of AD diagnostic data containing eye movements. Jang \textit{et. al} \cite{jang2021classification} utilized a machine learning-based deep classification neural network to extract features from multimodal eye movement indicators for AD diagnosis. Pereira \textit{et. al} \cite{pereira2020visual} also introduced machine learning to process eye movements collected in visual search tasks. In order to take full advantage of the deep learning network's ability to process different categories of data, Barral \textit{et. al} \cite{barral2020non} designed deep learning-based classifiers to extract eye movement features and language features respectively of subjects in a visual search task, and fused the two features for AD diagnosis. 

Although these experiments have demonstrated that deep learning can efficiently and intelligently process eye movement data and can be used in AD diagnosis, there is a relative lack of application of attention modules in the network architectures, even though attentional mechanisms have been shown to be effective in processing multimodal information.

As an important conception in deep learning field, attention neural network tends to focus on the distinctive parts when processing large amounts of information, which has been widely used in diverse application domains \cite{niu2021review}. The most commonly used domains for attention modules are natural language processing (NLP) \cite{bahdanau2014neural,liu2016neural,mi2016supervised} and computer vision (CV) \cite{mnih2014recurrent,jaderberg2015spatial,hu2018squeeze}. For example, local attention has been verified by Luong \textit{et. al} \cite{luong2015effective} to improve translation performance by focusing on a portion of the words in the source sentence at a time. Mnih \textit{et. al} \cite{mnih2014recurrent} presented a spatial attention model that is formulated as a single RNN that takes a glimpse window as its input and uses the internal state of the network to select the next location to focus on as well as to generate control signals in a dynamic environment.

To exploit the superiority of visual attention modules in eye movement analysis, we used the visual attention module in the free viewing task to extract visual saliency information between the visual stimulus and eye movements. We also utilized the serial attention module in the saliency comparison process due to the temporal nature of the eye movement data, which allowed the network to give more attention to abnormal eye movement behaviors.


Although a variety of eye movement indicators have been shown to be valid in the early diagnosis of AD, very limited studies have focused on the correlation between eye movements and visual stimuli in eye movement analyses. It has been demonstrated that the saliency distribution of the visual stimulus that elicits the eye movement behaviors affects the differentiation between AD patients and healthy controls \cite{tales2004effects}, so that a joint analysis of visual stimuli with the corresponding eye movements can lead to more objective diagnostic results. 

To overcome this drawback, we designed an image free-viewing task and jointly analyzed the saliency distribution of visual stimuli with the corresponding eye movements. We performed a comparative analysis based on the saliency of visual stimuli and eye movements, and made a diagnosis of AD based on the degree of their correlation. 

\section{Method}
\begin{figure*}
\centerline{\includegraphics[width=2.0\columnwidth]{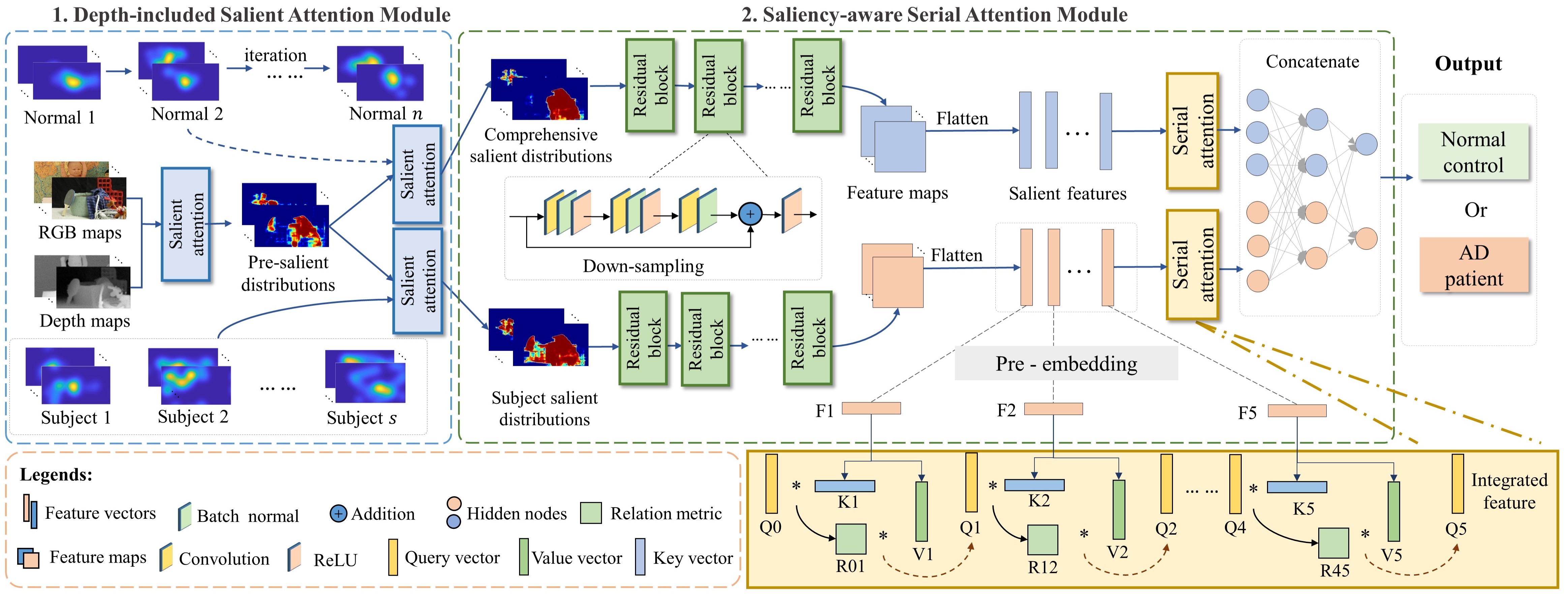}}
\caption{The overall structure of proposed DISCN with two parts: an integration module for fusing heatmaps and RGB-D visual stimuli, and a serial attention module for combining temporal visual saliency features.}
\label{DISCN}
\end{figure*}

In this section, we introduce our proposed DISCN in detail, including the notations and preliminaries, the mathematical descriptions of the two parts in DISCN: a depth-included salient attention module and a saliency-aware serial attention module as shown in Fig. \ref{DISCN}. The main idea is to jointly analyse the salient distributions of visual stimuli and the corresponding visual attention to detect the abnormal eye movements of AD patients caused by cognitive declines. To evaluate comprehensive salient distributions of visual stimuli, the RGB-D images and normal heatmaps are integrated with the salient attention module proposed in \cite{liu2021visual}. Meanwhile, a saliency-aware serial attention module is introduced to merge temporal features separately from salient distributions of visual stimuli and subjects. Finally, multi-layer perception (MLP) is used as a classifier, and the result of which is considered as a diagnosing result.

\subsection{Notations and preliminaries}
In the following subsections, we represent a set of 3-channel images with notation $\textbf{I}$ along with meaningful superscripts and subscripts, e.g., $\textbf{I}_{o}$ denotes a set of original images of visual stimuli. $\textbf{T}$ is defined as a set of tokens, which is composed of $n$ single tokens $\textbf{t}$, i.e., $\textbf{T}=[\textbf{t}^1, \textbf{t}^2, \dots, \textbf{t}^n]$. The feature maps calculated by residual layers are defined as $\textbf{F}$, and feature vectors are defined as $\textbf{v}$. Other matrices used in the paper are represented by uppercase bold letters, and vectors are represented by lowercase bold letters. The dimensions and meanings of all notations will be explained after it is defined. 

\subsection{Depth-included salient attention module for preliminary saliency map evaluation of visual stimuli}
The salient attention module is used for integrating original images and depth images of visual stimuli, as well as integrating visual stimuli and heatmaps that represent human visual attention with attention structures. The original images and depth images of $N$ visual stimuli are firstly integrated to evaluate preliminary saliency maps with a salient attention (SAA) block, which is constructed by "Image-to-Token" (I2T) layers, “Tokens-to-Token” (T2T) layers \cite{T2T2021}, transformer layers \cite{2017attentionisallyouneed}, "Cross-Modality Transformer" (CMT) layers, and "Token Attention" (TA) layers. The whole process of integrating original images and depth images of visual stimuli by SAA is defined as follows:

\begin{align} \mathcal{F}_{SAA}(\textbf{I}_{o}, \textbf{I}_d)=\left\{\begin{aligned}
&
\textbf{T}_o^0=\mathcal{F}_{I2T}(\textbf{I}_{o}),\\[6pt]
&
\textbf{T}_d^0=\mathcal{F}_{I2T}(\textbf{I}_{d}),\\[6pt]
&
\textbf{T}^i_{o}=\mathcal{F}_{T2T}(\textbf{T}_o^{i-1}),i=1,2,\\[6pt]
&
\textbf{T}^i_d=\mathcal{F}_{T2T}(\textbf{T}_d^{i-1}),,i=1,2,\\[6pt]
&
\textbf{T}^2_{od}=\mathcal{F}_{Cmt}(\textbf{T}^2_{o},\textbf{T}^2_{d}),\\[6pt]
&
\textbf{T}^i_{od}=\mathcal{F}_{Tr}(\mathcal{F}^{-1}_{T2T}(\textbf{T}^{i+1}_{od})+\textbf{T}^{i}_{od}),i=0,1,\\[6pt]
&
\textbf{T}_{od}^s=\mathcal{F}_{TA}(\textbf{T}^0_{od}),\\[6pt]
&
\textbf{I}_{pre}^{s}=\mathcal{F}_{T2T}^{-1}(\textbf{T}^s_{od}),\\[6pt]
 \end{aligned}\right.\end{align}where $\textbf{I}_{o},\textbf{I}_{d}\in\mathbb{R}^{N\times C\times H\times W}$ are the original images (RGB images) and depth images of the $N$ visual stimuli, which have $C$ channels and size of $W\times H$. $\textbf{T}_o^i,\textbf{T}_d^i,\textbf{T}_{od}^i\in \mathbb{R}^{N\times n_i\times d_i}$ and $\textbf{T}^s_{od}\in \mathbb{R}^{N\times n_0\times d_0}$ are the intermediate tokens in the saliency map evaluation process, where $n_i$ is the number of tokens of each visual stimulus, and $d_i$ is the dimension of each token. $\textbf{I}^s_{pre}\in \mathbb{R}^{N\times C\times H\times W}$ is the preliminary saliency maps evaluated from original and depth images, which have the same size. The function $\mathcal{F}_{Tr}(\cdot)$ represents the transformer structure \cite{2017attentionisallyouneed} of sequence-to-sequence mode that can model global dependencies among all pixels in the whole image. The main functions and the variables are explained as follows:

\begin{itemize}
    \item \textbf{Image to Tokens $\mathcal{F}_{I2T}(\cdot)$}

\begin{figure}[t]
\centerline{\includegraphics[width=1.02\columnwidth]{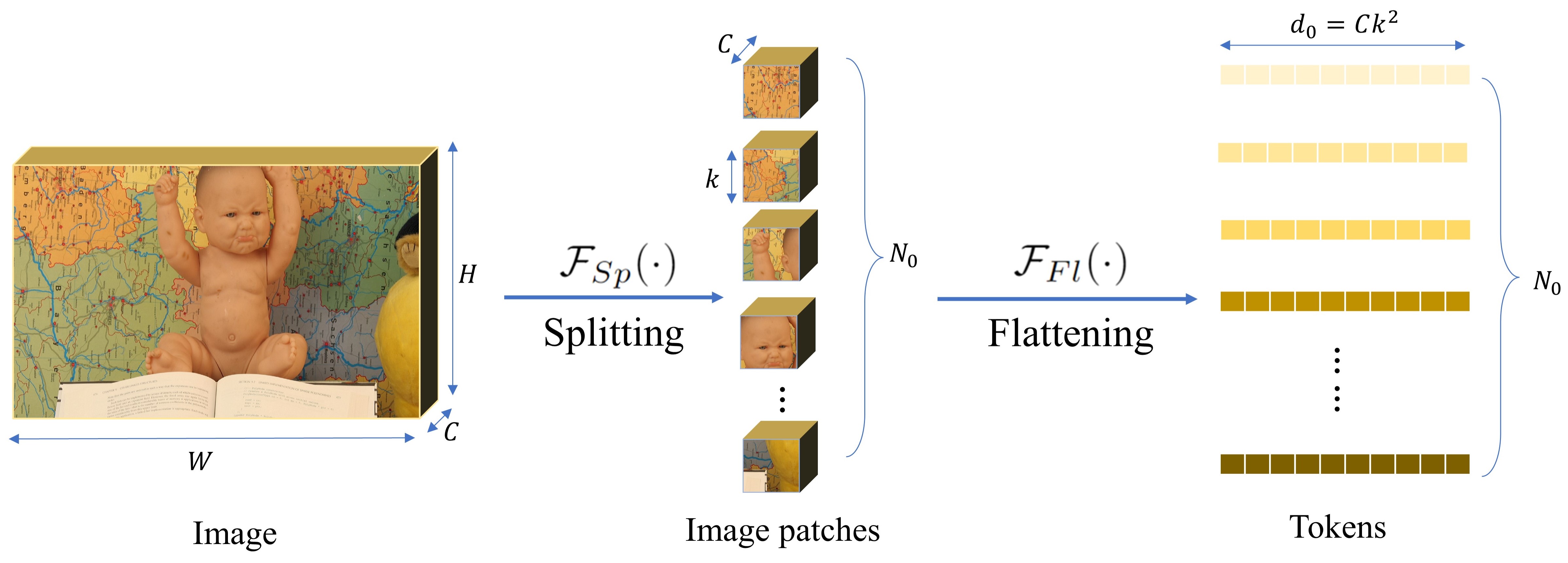}}
\caption{The process of I2T that consists of a splitting step and a flattening step for converting an image into tokens.}
\vspace{-1em}
\label{split function}
\end{figure}

$\mathcal{F}_{I2T}(\cdot)$ denotes the operation that converts an image into a set of vectors i.e. tokens, which includes two steps: splitting and flattening, as shown in Fig. \ref{split function}. 

\begin{equation}
    \mathcal{F}_{I2T}(\cdot)=\mathcal{F}_{Sp}(\mathcal{F}_{Fl}(\cdot)),
\end{equation}where $\mathcal{F}_{Sp}(\cdot)$ denotes the operation of splitting images into image patches, and $\mathcal{F}_{Fl}{(\cdot)}$ is the operation of flattening image patches by pixel.

\item \textbf{Tokens to Token $\mathcal{F}_{T2T}(\cdot)$}

\begin{figure}[t]
\centerline{\includegraphics[width=1.02\columnwidth]{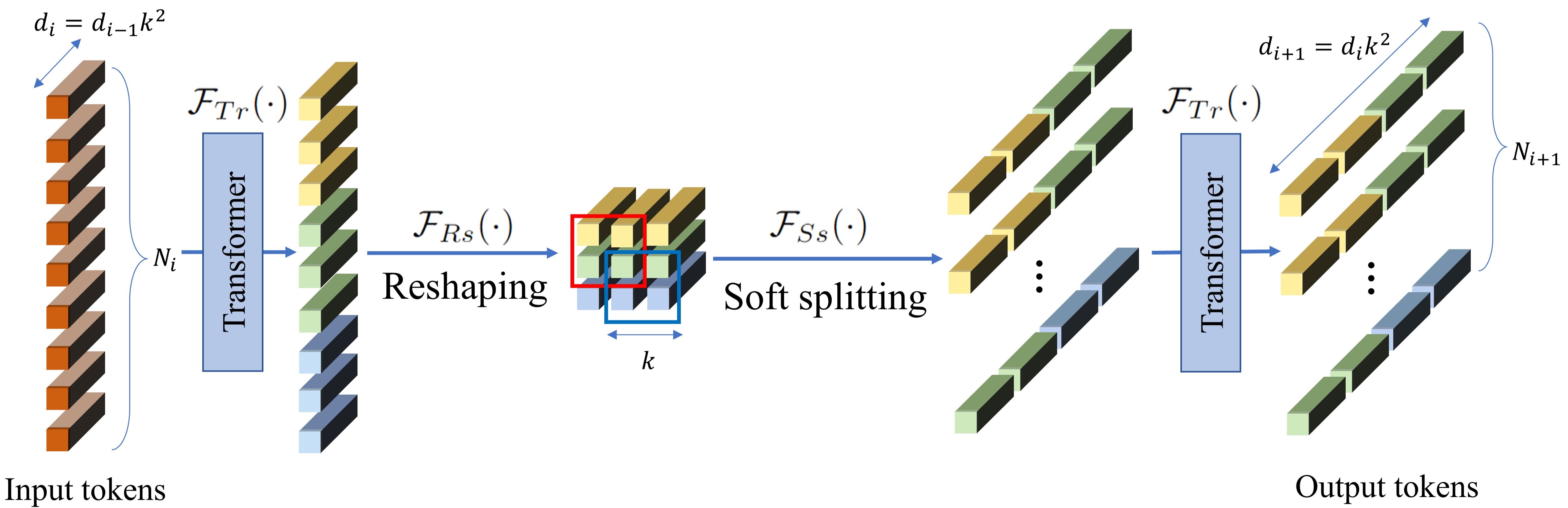}}
\caption{The process of T2T that consists of transformers, a reshaping step and a soft-split step for converting input tokens into high-dimensional tokens.}
\vspace{-1em}
\label{T2T}
\end{figure}

$\mathcal{F}_{T2T}(\cdot)$ denotes the operation that convert tokens into high-dimensional tokens with two transformers, a reshaping step and a soft-split step, as shown in Fig \ref{T2T}. $\mathcal{F}_{T2T}(\cdot)$ is defined as follows:

\begin{equation}
    \mathcal{F}_{T2T}(\cdot)=\mathcal{F}_{Tr}(\mathcal{F}_{Rs}(\mathcal{F}_{Ss}(\mathcal{F}_{Tr}(\cdot)))),
\end{equation}where $\mathcal{F}_{Rs}(\cdot)$ is the reshaping step that combines tokens into metrics, and $\mathcal{F}_{Ss}$ is the soft splitting function that splits the high-dimensional images into new vector sets in a manner like convolution. The window size of each split is $k\times k$ with a padding of $p$ pixels and a stride of $s$. Pixels in each window are connected by the channel dimension to get high-dimensional tokens. The dimensions of tokens are converted from $N\times n_i\times d_i$ into $N\times n_{i+1}\times d_{i+1}$ in the $i$th T2T layer, where $n_{i+1}=(\sqrt{n_{i}}+2p-k)/{s}+1$ and $d_{i+1}=d_{i}\times k^2$. Then a transformer is conducted to the high-dimensional tokens and finally get the output tokens $\textbf{T}_{o}^{i+1},\textbf{T}_d^{i+1}\in \mathbb{R}^{N\times n_{i+1}\times d_{i+1}}$, where $i=1,2$ is the number of T2T layers.

\item \textbf{Cross-Modality Transformer} $\mathcal{F}_{Cmt}(\cdot)$

A CMT layer is adopted to combine the two set of tokens from original and depth images separately to get cross-modality tokens $\textbf{T}_{od}^2$. Function $\mathcal{F}_{Cmt}(\cdot)$ as follows:

\begin{small}
\begin{align}
\mathcal{F}_{Cmt}(\textbf{T}_o^{2},\textbf{T}_d^{2})=\left\{\begin{aligned}
&\textbf{Q}_{o},\textbf{Q}_{d}=\textbf{W}_{o}^q \textbf{T}_o^{2},\textbf{W}_{d}^q\textbf{T}_d^{2},\\[6pt]
&\textbf{K}_{o},\textbf{K}_{d}=\textbf{W}_{o}^k\textbf{T}_d^{2},\textbf{W}_{d}^k\textbf{T}_d^{2},\\[6pt]
&\textbf{V}_{o},\textbf{V}_{d}=\textbf{W}_{o}^v\textbf{T}_o^{2},\textbf{W}_{d}^v\textbf{T}_d^{2},\\[6pt]
&\textbf{T}_o^{c}=Softmax(\textbf{Q}_{o}(\textbf{K}_{d})^T)\textbf{V}_{d},\\[6pt]
&\textbf{T}_d^{c}=Softmax(\textbf{Q}_{d}(\textbf{K}_{o})^T)\textbf{V}_{o},\\[6pt]
&\textbf{T}_{od}^2= \textbf{W}^{c}([\textbf{T}_o^{c},\textbf{T}_d^{c}])
\end{aligned}\right.
\label{I2T}
\end{align}
\end{small}where $\textbf{Q}_{o},\textbf{K}_{o},\textbf{V}_{o},\textbf{Q}_{d},\textbf{K}_{d},\textbf{V}_{d}\in \mathbb{R}^{N\times n_2\times d_2}$ are the query, key and value vectors of original and depth images, $\textbf{W}_{o}^q,\textbf{W}_{d}^q,\textbf{W}_{o}^k,\textbf{W}_{d}^k,\textbf{W}_{o}^v,\textbf{W}_{d}^v\in \mathbb{R}^{n\times d_2\times d_2}$ are learnable matrices for linear embedding from tokens $\textbf{T}_o^{2},\textbf{T}_d^{2}$ into query, key, and value vectors. $\textbf{T}_o^{c},\textbf{T}_d^{o}\in \mathbb{R}^{N\times n_2\times d_2}$ are tokens that combines information from both original images and depth images. Then the two output tokens are concatenated and projected linearly into a token set $\textbf{T}_{od}^2\in \mathbb{R}^{N\times n_2\times d_2}$. $\textbf{W}^c\in \mathbb{R}^{N\times n_2\times 2n_2}$ is a learnable matrix for linear projection, and $[,]$ denotes concatenation by the token number dimension $n_2$.

\item \textbf{Reversal Tokens to Token} $\mathcal{F}_{T2T}^{-1}(\cdot)$

\begin{figure}[t]
\centerline{\includegraphics[width=1.02\columnwidth]{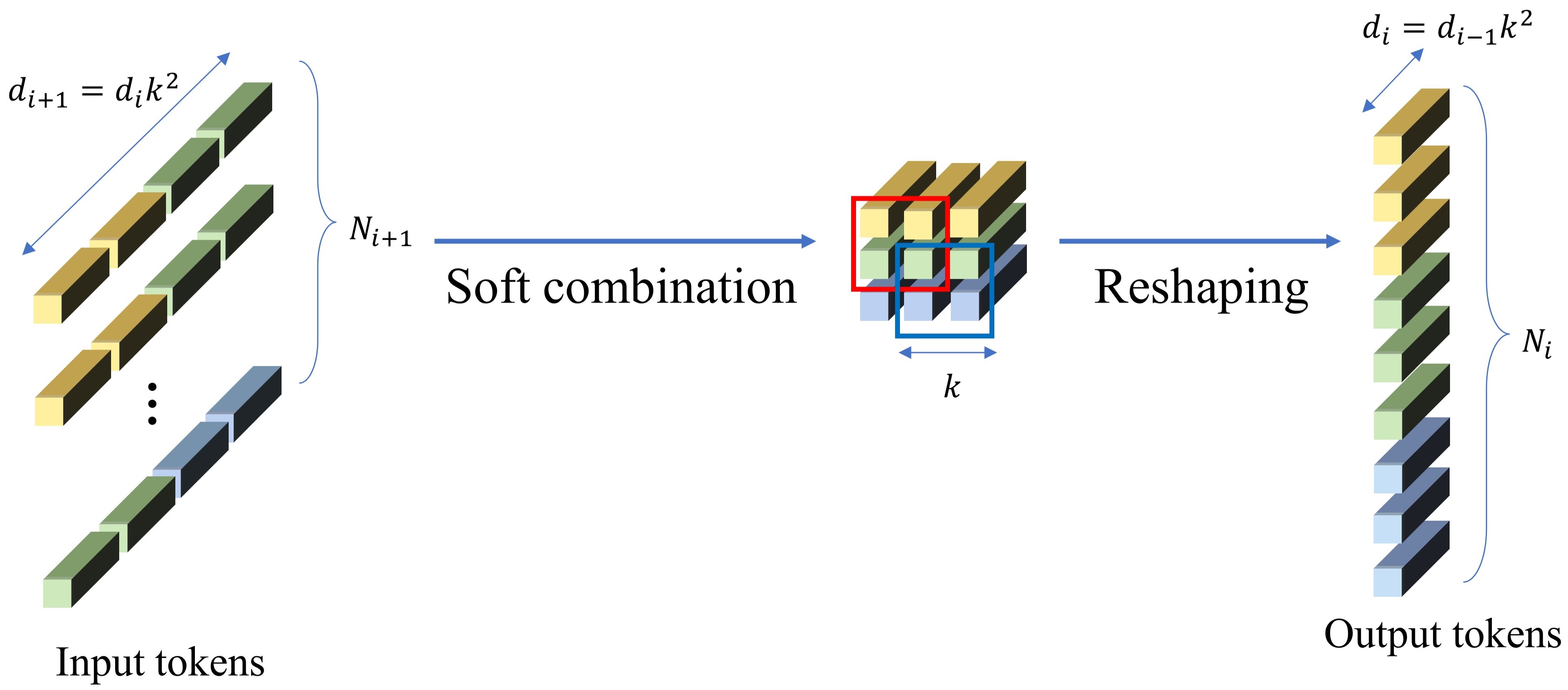}}
\caption{The process of reversal T2T that consists of a soft-combine step and a reshaping step for converting high-dimensional tokens into more lower-dimensional tokens.}
\vspace{-1em}
\label{RT2T}
\end{figure}

The reversal tokens to token operation is a reversal process of T2T, and is represented with $\mathcal{F}^{-1}_{T2T}(\cdot)$. Reversal T2T aims to recover the high-dimensional tokens into their original shapes. As shown in Fig. \ref{RT2T}, $\mathcal{F}^{-1}_{T2T}(\cdot)$ is defined as follows:

\begin{equation}
    \mathcal{F}_{T2T}^{-1}(\cdot)=\mathcal{F}_{Tr}(\mathcal{F}_{Ss}^{-1}(\mathcal{F}_{Rs}^{-1}(\cdot))),
\end{equation}where $\mathcal{F}_{Ss}^{-1}(\cdot)$ is the reversal process of soft splitting, and $\mathcal{F}_{Rs}^{-1}(\cdot)$ is the reversal process of reshaping. Reversal T2T is applied on $T_{od}^{3}$ twice in a residual manner:

\begin{equation}
\textbf{T}_{od}^i=\mathcal{F}_{T2T}^{-1}(\textbf{T}_{od}^{i+1})+\textbf{T}_o^{i}
\end{equation}where $\textbf{T}^i_{od}\in \mathbb{R}^{N\times n_i\times d_i},i=0,1$ are the outputs of each reversal T2T operation. Among them, $\textbf{T}^0_{od}=[\textbf{t}_s,\textbf{t}_1,\dots,\textbf{t}_{n_0-1}]$, where the first token $\textbf{t}_s$ is seen as the salient token to learn the salient information from the whole token set $\textbf{T}^0_{od}$. 

\item \textbf{Token Attention} $\mathcal{F}_{TA}(\cdot)$

A Token attention (TA) mechanism is adopted for saliency evaluation, which is defined as follows:

\begin{small}
\begin{align}
\mathcal{F}_{TA}(\textbf{T}^C_0)=\left\{\begin{aligned}
&\textbf{Q}_s=\textbf{W}_s^q\textbf{t}_s,\\[6pt]
&\textbf{K}_s=\textbf{W}_s^k\textbf{t}_s,\\[6pt]
&\textbf{V}_s=\textbf{W}_s^v\textbf{t}_s,\\[6pt]
&\textbf{Q}_i=\textbf{W}_i^q\textbf{t}_i,i=1,2,\cdots,n_0-1\\[6pt]
&\hat{\textbf{t}}_{s}=Softmax(\textbf{Q}_{s}\textbf{K}_{s}^T)\textbf{V}_s,\\[6pt]
&\hat{\textbf{t}}_{i}=Softmax(\textbf{Q}_{i}\textbf{K}_{s}^T)\textbf{V}_s, i=1,\dots,n_0-1,\\[6pt]
&\textbf{T}_{od}^s=[\hat{\textbf{t}}_{s},\hat{\textbf{t}}_{1},\dots,\hat{\textbf{t}}_{n_0-1}],\\[6pt]
\end{aligned}\right.
\label{RT2T}
\end{align}
\end{small}where $\textbf{t}_s,\textbf{t}_1,\dots,\textbf{t}_{n_0-1}\in \mathbb{R}^{N\times 1\times d_0}$ are the tokens of token set $\textbf{T}^0_{od}$, $\textbf{Q}_s,\textbf{K}_s,\textbf{V}_s\in \mathbb{R}^{N\times 1\times d_0}$ are the query,key and value vector of salient token $\textbf{t}_s$. $\textbf{W}_s^q,\textbf{W}_s^k,\textbf{W}_s^v\in \mathbb{R}^{N\times d_0\times d_0}$ are learnable matrices for linear projection. $\textbf{Q}_1,\textbf{Q}_2,\dots,\textbf{Q}_{n_0-1}\in \mathbb{R}^{N\times 1\times d_0}$ are the query vectors of the rest tokens, and $\textbf{W}_1^q,\textbf{W}_2^q,\dots,\textbf{W}_{n_0-1}^q\in \mathbb{R}^{n\times d_0\times d_0}$ are learnable matrices for linear projection. The output tokens $\hat{\textbf{t}}_s$ and $\hat{\textbf{t}}_i,i=1,2,\dots,n_0-1$ compose the salient token set $\textbf{T}_{od}^s\in \mathbb{R}^{N\times n_0\times d_0}$. The salient token set is reshaped into a preliminary saliency map $\textbf{I}^{s}_{pre}$ by a reversal T2T function.
\end{itemize}

\subsection{Salient attention module for integrating visual stimuli with human visual attention}

The human visual attention can be represented by heatmaps of the same shape with visual stimuli. Heatmaps are divided into two sets: a normal set $\mathbb{D}_{NC}$ and a subject set $\mathbb{D}_{SU}$. The $\mathbb{D}_{NC}$ contains heatmaps of $n$ normal controls. $\mathbb{D}_{SU}$ contains heatmaps of $s$ subjects including 1:1 normal controls and AD patients that need to be classified. The definitions of $\mathbb{D}_{NC}$ and $\mathbb{D}_{SU}$ are as follows:

\begin{small}
\begin{align}
\left\{\begin{aligned}
&\mathbb{D}_{NC}=\{\textbf{H}^1_{NC},\textbf{H}^2_{NC},\dots,\textbf{H}^{n}_{NC}\},\\[6pt]
&\mathbb{D}_{SU}=\{\textbf{H}^1_{SU},\textbf{H}^2_{SU},\dots,\textbf{H}^{s}_{SU}\},\\[6pt]
\end{aligned}\right.
\label{heatmap sets}
\end{align}
\end{small}where $\textbf{H}^i_{NC},\textbf{H}^j_{SU}\in \mathbb{R}^{N\times C\times H\times W},i=1,2,\dots,n,j=1,2,\dots,s$ are the heatmaps corresponding to $N$ visual stimuli. For the subject set, preliminary saliency maps are integrated with heatmaps of each individuals with SAA module mentioned in Section III-B:

\begin{equation}
    \textbf{I}^i_{sub}=\mathcal{F}_{SAA}(\textbf{I}_{pre}^{s},\textbf{H}^i_{SU}),i=1,2,\dots,s,
\end{equation}where $\textbf{I}^i_{sub}\in \mathbb{R}^{N\times C\times H\times W}$ repersents the saliency maps of the $i$th subject containing salient information of both visual stimuli and human visual attention. As for the normal control set, heatmaps are integrated with preliminary saliency maps in an iterative manner:

\begin{small}
\begin{align}
\left\{\begin{aligned}
&\textbf{I}^s_{i+1}=\mathcal{F}_{SAA}(\textbf{I}_{pre}^{s},\textbf{H}_{NC}^{i+1}),i=0,1,\dots,n-1,\\[6pt]
&\textbf{I}^{s}_{0}=\textbf{I}^{s}_{pre},\\[6pt]
&\textbf{I}_{com}^{s}=\textbf{I}_{n}^{s},
\end{aligned}\right.
\label{heatmap sets}
\end{align}
\end{small}where $\textbf{I}^{s}_{com}\in \mathbb{R}^{N\times C\times H\times W}$ denotes the comprehensive saliency maps that includes visual salient information of both RGB-D visual stimuli and $n$ normal controls. $\textbf{I}^s_i,i=1,2,\dots,n-1$ are the intermediate images of the $i$th iteration. 

\subsection{Saliency-aware serial attention module for feature fusion}

The saliency-aware serial attention module will extract visual saliency-aware features from the two sets of saliency maps, and fuse the features with serial attention module.

Given the $i$th subject saliency maps $\textbf{I}^{i}_{sub}$ and the comprehensive saliency maps $\textbf{I}^{s}_{com}$, $r$ residual blocks are implemented on the two sets of saliency maps separately in cascade. The $j$th residual block is defined as function $\mathcal{F}_{Res}^{j}(\cdot)$:

\begin{small}
\begin{align}
\mathcal{F}_{Res}^{j}(\textbf{I}^{s}_{j-1})\left\{\begin{aligned}
&\textbf{F}^{1}_j=ReLU(BN(Conv(\textbf{I}^{s}_{j-1}))),\\[6pt]
&\textbf{F}^{2}_j=ReLU(BN(Conv(\textbf{F}^{1}_j))),\\[6pt]
&\textbf{F}^{3}_j=BN(Conv(\textbf{F}^{2}_j)),\\[6pt]
&\textbf{I}^{s}_{j}=ReLU(\mathcal{F}_{US}(\textbf{I}^{s}_{j-1})+\textbf{F}^{3}_j),
\end{aligned}\right.
\label{residual}
\end{align}
\end{small}where $\textbf{I}^{s}_{j-1}\in \mathbb{R}^{N\times C_{j-1}\times H_{j-1}\times W_{j-1}}$ are the outputs of the $(j-1)$th residual block, where $j=1,2,\dots,r$ and $\textbf{I}^{s}_{0}=\textbf{I}^{s}_{com}$. $Conv(\cdot)$ denotes a convolutional layer, $BN(\cdot)$ denotes a batch normalization layer and $ReLU(\cdot)$ denotes a ReLU activation layer. $\textbf{F}^i_j\in \mathbb{R}^{N\times C_j\times H_j\times W_j},i=1,2,3$ are the feature maps of $j$th residual block. $\mathcal{F}_{US}(\cdot)$ indicates up-sampling operation on the $C_j$ dimension. $r$ residual blocks are applied on $\textbf{I}^{s}_{com}$ and $\textbf{I}^{i}_{sub}$, and get feature maps of the $r$th residual layer, which are flattened into vector sets $\textbf{V}_{com},\textbf{V}_{sub}^{i}\in \mathbb{R}^{N\times d_0}$, where $d_0=C_{r}\times H_{r}\times W_{r}$. $\textbf{V}_{com}=[\textbf{v}^{1}_{com},\textbf{v}^{2}_{com},\dots\textbf{v}^{n}_{com}]$ and $\textbf{V}^{i}_{sub}=[\textbf{v}_{i}^1,\textbf{v}_{i}^2,\dots,\textbf{v}_{i}^n]$ are the serial features in temporal order of comprehensive and subject saliency maps respectively, where $\textbf{v}_{com}^j,\textbf{v}^{j}_i\in \mathbb{R}^{d_0},j=1,2,\dots,n$. The serial features are preliminarily embedded into feature serials $[\textbf{f}^{1}_{com},\textbf{f}^{2}_{com},\dots,\textbf{f}^{n}_{com}]$ and $[\textbf{f}^{1}_{i},\textbf{f}^{2}_{i},\dots,\textbf{f}^{n}_{i}]$, where $\textbf{f}^{j}_{com},\textbf{f}^{j}_{i}\in \mathbb{R}^{d_1},j=1,2,\dots,n$:

\begin{small}
\begin{align}
\left\{\begin{aligned}
&\textbf{f}^{j}_{com}=\textbf{W}_{com}\textbf{v}^{j}_{com},\\[6pt]
&\textbf{f}^{j}_{i}=\textbf{W}_{sub}\textbf{v}^{j}_{sub},
\end{aligned}\right.
\label{residual}
\end{align}
\end{small}where $\textbf{W}_{com},\textbf{W}_{sub}\in \mathbb{R}^{d_0\times d_1}$ is matrices used for preliminary embedding. Then the feature serials are embedded into key and value vectors:

\begin{small}
\begin{align}
\left\{\begin{aligned}
&\textbf{k}_{com}^j=\textbf{W}_{com}^k\textbf{f}_{com}^j,\\[6pt]
&\textbf{v}_{com}^j=\textbf{W}_{com}^v\textbf{f}_{com}^j,\\[6pt]
&\textbf{k}_{sub}^j=\textbf{W}_{sub}^k\textbf{f}^{j}_{sub},\\[6pt]
&\textbf{v}_{sub}^j=\textbf{W}_{sub}^v\textbf{f}^{j}_{sub},
\end{aligned}\right.
\end{align}
\end{small}where $\textbf{W}_{com}^k,\textbf{W}_{com}^v,\textbf{W}_{sub}^k,\textbf{W}_{sub}^v\in \mathbb{R}^{d_1\times d_2}$ are matrices for prelimibary embedding. $\textbf{k}_{com}^j,\textbf{v}_{com}^j,\textbf{k}_{sub}^j,\textbf{v}_{sub}^j\in \mathbb{R}^{d_2},j=1,2,\dots,n$ are keys and values of the $j$th vector in the feature serial. The vectors in feature serial are integrated by serial attention module:

\begin{small}
\begin{align}
\left\{\begin{aligned}
&\textbf{q}^{j}_{com}=Softmax(\textbf{q}_{com}^{j-1} \textbf{k}_{com}^j) \textbf{v}_{com}^j,\\[6pt]
&\textbf{q}^{j}_{sub}=Softmax(\textbf{q}^{j-1}_{sub} \textbf{k}^{j}_{sub})\textbf{v}^{j}_{sub},
\end{aligned}\right.
\end{align}
\end{small}where $j=1,2,\dots,n$ and $\textbf{q}_{com}^0,\textbf{q}_{com}^0\in \mathbb{R}^{d_2}$ are learnable parameters. $\textbf{q}_{com}^j$ and $\textbf{q}_{sub}^j$ are calculated by previous features $\textbf{f}_{com}^{j-1}$ and $\textbf{f}_{sub}^{j-1}$, which presents the temporal order of the serial. $\textbf{q}_{com}^n$ and $\textbf{q}_{sub}^n$ are concatenated into vector $\textbf{q}^n$:

\begin{equation}
    \textbf{q}^n=[\textbf{q}_{com}^n,\textbf{q}_{sub}^n],
\end{equation}where $\textbf{q}^n\in \mathbb{R}^{d_3},d_3=2\times d_2$ is the final vector in serial attention, which is inputted into a 2-layered MLP:

\begin{equation}
\textbf{q}^n_2=Softmax(\textbf{W}_2\times ReLU(\textbf{W}_1\times \textbf{q}^n)),
\end{equation}where $\textbf{W}_1\in \mathbb{R}^{d_3\times d_4}$, $\textbf{W}_2\in \mathbb{R}^{d_4\times 2}$ are linear matrices, and $\textbf{q}^n_2\in \mathbb{R}^{2}$ is the classification result, whose first item $q$ represents the probability that the $i$th subject is diagnosed as an AD patient. A binary cross entropy loss (BCE) is applied to calculate the classification loss $L$ by $q$ and the label of $i$th subject $t$:

\begin{equation}
    L=\mathcal{F}_{BCE}(q,t)=-tlog(q)+(1-t)log(1-q),
\end{equation}

In the convergence procedure, all the learnable parameters of DISCN are updated by backward gradient descent, which propagates the gradient information of each parameter through the entire DISCN and optimizes all the parameters according to the fitting loss. Parameters could be updated in each training ergodic of all items in the training dataset:

\begin{equation}
    \mathcal{W}=\{w_0,w_1,\dots,w_i,\dots\}=\mathcal{W}-\alpha\frac{\partial{L}}{\partial{\mathcal{W}}},
    \label{eq9}
\end{equation}where $\mathcal{W}$ denotes all parameters and $\alpha$ decides the rate of parameters updating, namely learning rate. With enough epoches, all parameters will be updated until the loss function reaches its stable and minimum value, which indices that the proposed model is at its optimum state of best-fit the predictions to labels.

\section{Experiments}
In this section, we introduce the basic information of participants in our projection and the eye movements collection process. We also visualize the process of comprehensive visual saliency evaluation. The designs and results of comparison experiments and ablation experiments with implementation details are presented in this section.

\subsection{Participants}

A total of 150 participants including 50 early AD patients and 100 normal controls are recruited in our study. The AD patients (23 males and 27 females) were aged from 54 to 84 years old who were recruited form the cognitive impairment clinics of Huanhu Hospital in Tianjin, China (\href{http://www.tnsi.org/}{http://www.tnsi.org/}). All AD patients met NINCDS-ADRDA criteria \cite{mckhann1984clinical}. In AD patients, humans with uncorrected dysfunctions of vision or hearing loss, mental disorders, or other symptoms that made them unable to complete the proposed free viewing task. The normal controls were recruited from age matched friends and relatives of patients who had no subjective or informant-based complaints of cognitive decline. The use of eye-tracking data for analysis of AD patients’
cognitive decline obeyed the World Medical Association Declaration of Helsinki \cite{world2001world}.  

As shown in Section III-C, we divide the 150 participants into a “subject set” $\mathbb{D}_{SU}$ and a “normal set” $\mathbb{D}_{NC}$. $\mathbb{D}_{NC}$ is composed with 50 normal controls, and the remaining 50 AD patients and 50 normal controls form $\mathbb{D}_{SU}$, namely $N_n=50$, $N_s=100$. Heatmaps in $\mathbb{D}_{NC}$ are used for iterative fusion with the visual stimuli, and participants in $\mathbb{D}_{SU}$ need to be diagnosed. $\mathbb{D}_{SU}$ is divided equally into five subsets to implement 5-fold cross-validation, as shown in TABLE I.

\begin{table}[htbp]
	\centering
    \label{split}
    \small
    \vspace{-1em}
    \tabcolsep=0.2cm
	\caption{The division of the subject set.}
	\begin{tabular}{ccccc}
		\toprule  
		 & Training set & Testing set & Sum & Label\\ 
        \midrule  
		AD patients & 40 & 10 & 50 & 0\\
        Normal controls & 40 & 10 & 50 & 1\\
        Sum & 80 & 20 & 100 & 0/1\\
		\bottomrule  
    \vspace{-2em}
	\end{tabular}
\end{table}






\subsection{Eye movements collection}

To collect eye movements of the participants under stereo visual stimuli, we perform a free viewing trial with a noninvasive binocular eye tracker. The eye movements collected from the free viewing trial are presented in the form of heatmaps.

\subsubsection{Stereo visual stimuli designs}

\begin{figure}[t]
\centerline{\includegraphics[width=1.02\columnwidth]{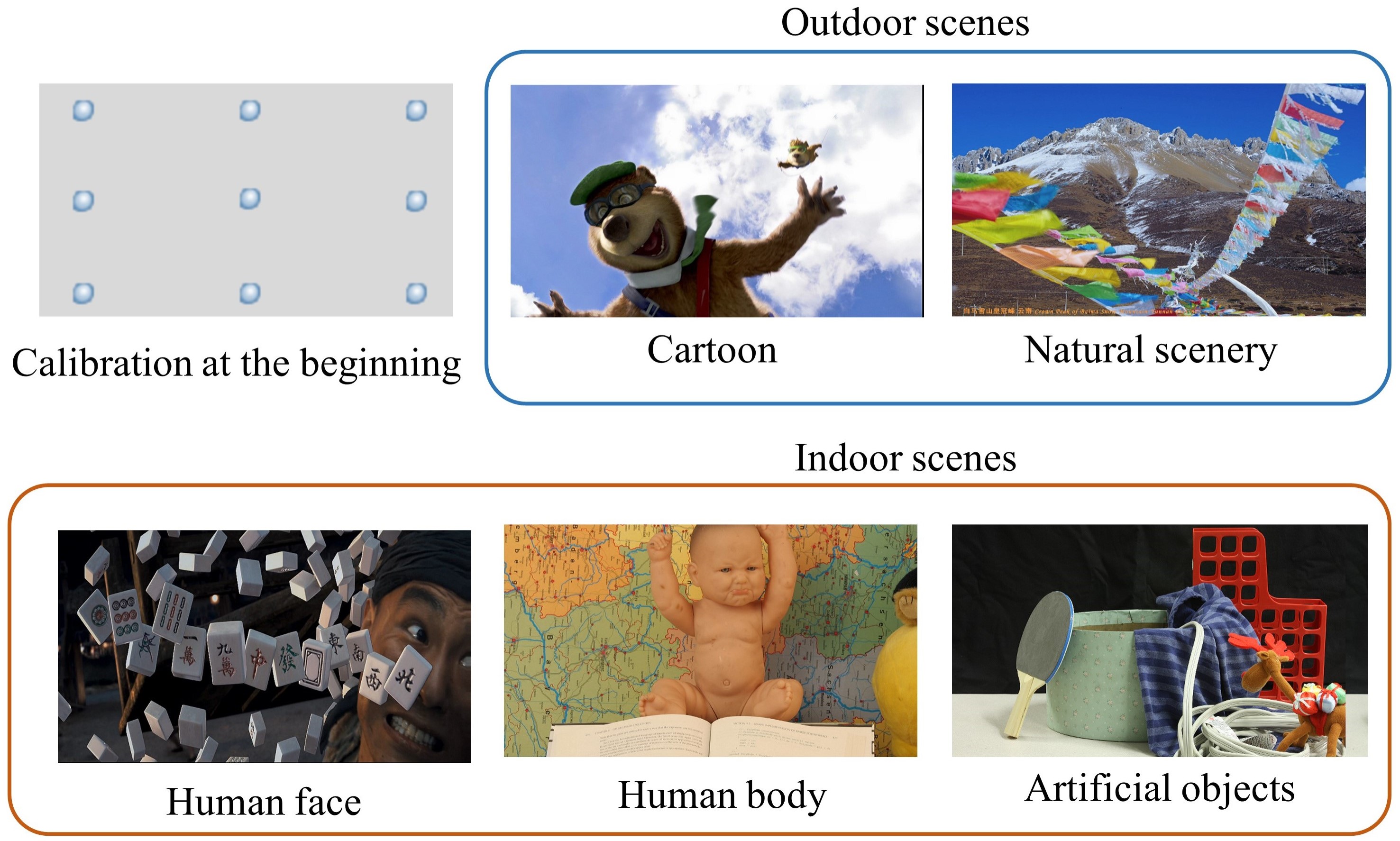}}
\caption{The stereo image stimuli of diverse styles presented to the subjects.}
\vspace{-1em}
\label{stimuli}
\vspace{0em}
\end{figure}

The stereo visual stimuli is a video including a calibration procedure at the beginning and a serial of images. The calibration procedure was applied to the participants for a more accurate eye movement estimation. After that, five images were displayed for 5 seconds through binocular stereo vision with a 1-second gap between each two of them. The contents of the stimuli contained indoor scenes, outdoor scenes with different styles including natural scenery, artificial objects, cartoon, human face and human body, which is presented in Fig. \ref{stimuli}. The depth maps corresponding to each RGB visual stimulus were achieved by calculating the disparities between the images of left-eye and right-eye views. 

The whole display of visual stimuli only took less than 1 minute for each participant, which satisfies the requirements of efficiency for large-scale AD diagnosis screening. The design of binocular vision mode is for introducing depth to stimulate richer brain and eye movement activities \cite{gonzalez2013breaking}. Diverse visual stimuli can provide subjects with different viewing experiences and stimulate rich eye movements to the maximize extent.

\subsubsection{Noninvasive eye tracker}

\begin{figure}[t]
\centerline{\includegraphics[width=1.02\columnwidth]{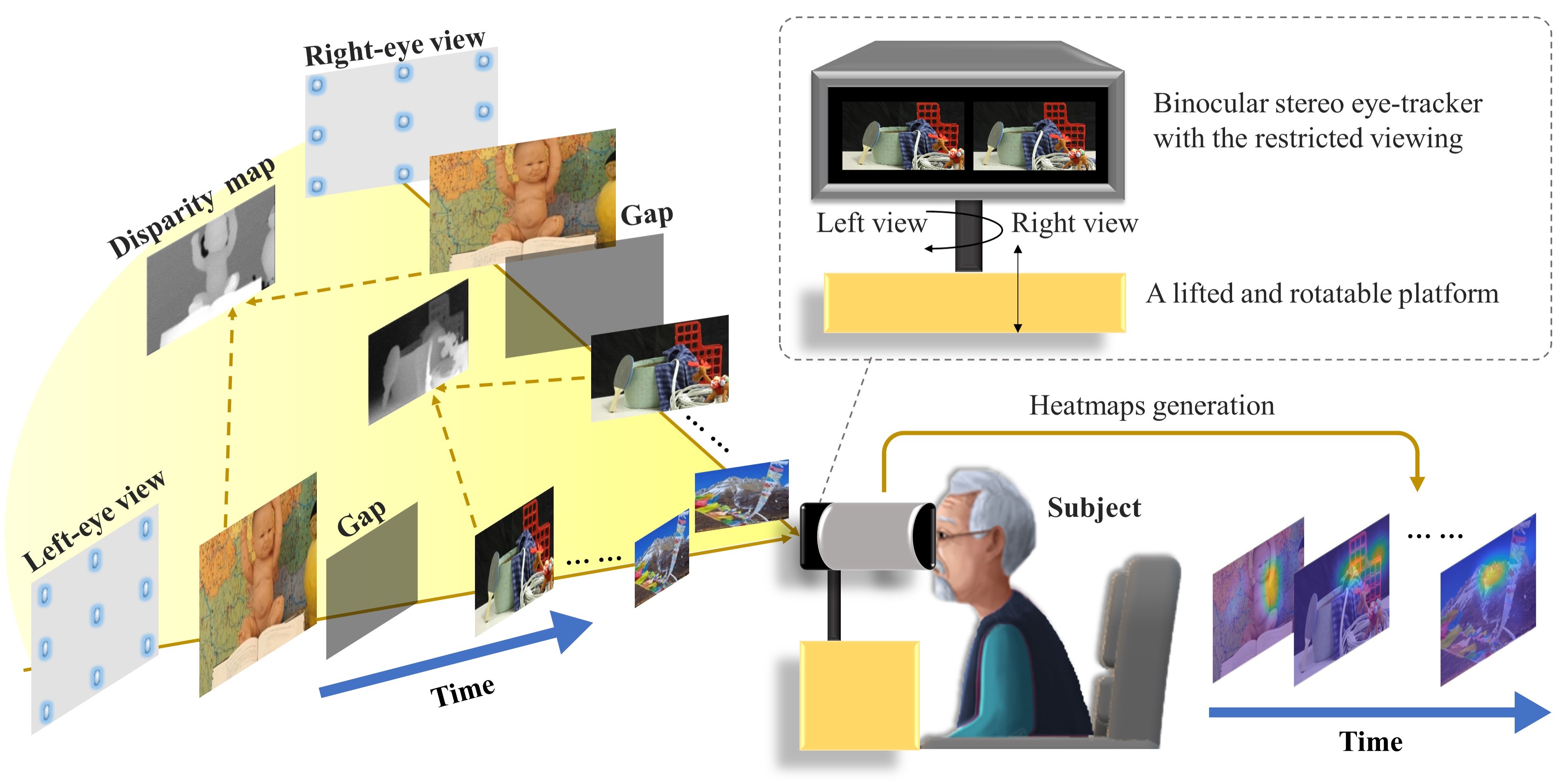}}
\caption{The process of eye movements collection and heatmaps generation.}
\vspace{-1em}
\label{process of data collection}
\end{figure}

During the free viewing task, eye movements of participants are recorded using an eye tracker with the stereo stimuli designed by Sun et al. \cite{sun2023}, this eye tracker estimates gaze positions at an average error of 1.85cm/0.15m over the workspace volume 2.4m$\times$4.0m$\times$7.9m. It displays stereo visual stimuli in the resolution of 1920$\times$1080@120 Hz in a limited vision without requiring the viewer to wear any accessories, which is friendly to participants, especially to the elders. The eye tracker can be adjusted in a 360-degree direction freely to meet the need of different users. It allows participants to experience a 3D immersive viewing by displaying the images of left-eye and right-eye perspectives separately through two sets of lenses. The eye movements are also collected in binocular vision by the inside cameras.

\subsubsection{Heatmaps of visual attention}

\begin{figure}[t]
\centerline{\includegraphics[width=1.02\columnwidth]{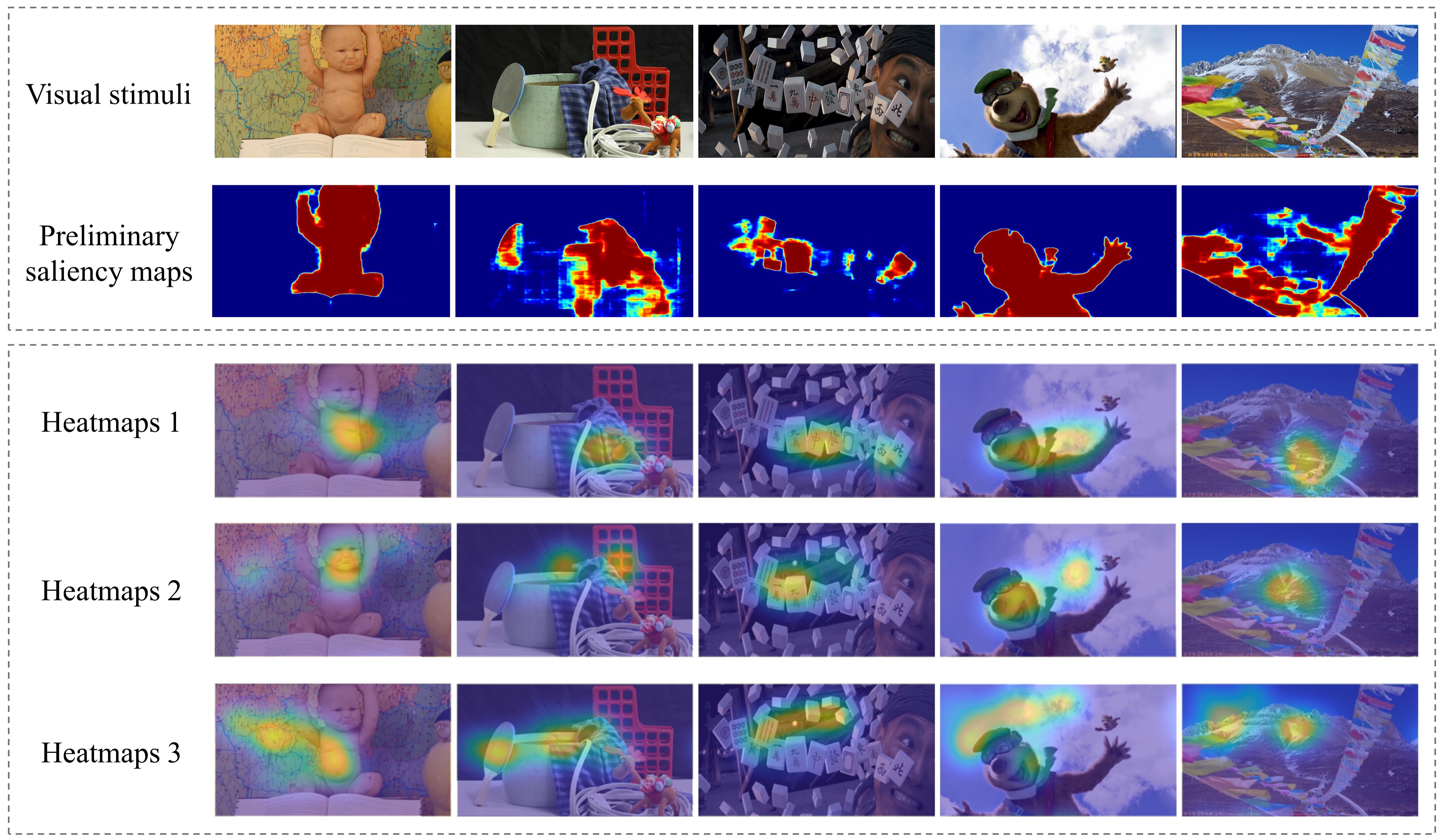}}
\caption{Several heatmaps that contain human visual attention information and the corresponding visual stimuli.}
\vspace{-1em}
\label{heatmap}
\end{figure}

The collected eye movements are transformed into heatmaps by iMap \cite{caldara2011map}, which reflect the distributions of visual attention. The generated heatmaps are the inputs of the proposed DISCN together with the RGB maps and depth maps of corresponding visual stimuli.
The heatmaps are evaluated with gaze positions and duration of each individual. In the heatmaps, color represents the duration-weighted spatial density of human gaze points, which reflects the visual attention level. The warmer color a region shows, the more visual attention an individual pays to it, as shown in Fig. \ref{heatmap}.

\subsection{Experiment designs}

\subsubsection{Ablation experiments}
The effectiveness of four modules in DISCN are verified, including depth integration, normals integration, residual feature extraction, and feature fusion. For depth integration and normals integration, we first visualize the process of evaluating comprehensive salient distributions for qualitative analysis. Then two models named noDEP and noNOR are designed to compare with DICSN. As for residual feature extraction and feature fusion, we further design corresponding models named noRES and noSEA for ablation. The models are introduced as follows:

\begin{itemize}
    \item noDEP: Depth affects the visual perception of subjects. In noDEP, original maps is integrated with normals directly by only one SAA.
    \item noNOR: Healthy visual attention is combined with pre-salient maps in DISCN. noNOR take only pre-salient distributions to compare with subject salient distributions.
    \item noRES: Residual blocks contain hierarchical residual structures that can address a common degeneration problem. In noRES,  residual blocks are replaced with traditional CNN layers with the same depth.
    \item noSEA: Serial attention module (SEA) fuses serial vectors in order, that models the temporal variations of salient features according to the order of visual stimuli. In noSEA, SEA is replaced by MLP layers.
\end{itemize}

Moreover, four feature fusing modules including MLP, GRU, LSTM and SEA are applied in DISCN to verify the effects. Among them, MLP fuses salient features with pure fully-connections and activation layers without temporal orders; GRU and LSTM are recurrent structures controled by gates, which receive features in order; SEA fuses salient features with both temporal orders and attention mechanism.

\subsubsection{Feature extraction structures}
To further explore the effects of different feature extraction and fusion module, three different feature extracting mechanisms including pure convolution (CNN), multi-scaled feature complementary (MSBPNet), and residual blocks (DISCN) are also tested based on the four feature fusing methods. The results validate the effectiveness of the network structure adopted by the DISCN from two dimensions: feature extraction and fusion.

\subsubsection{Comparison experiments}
In this part, the performance of DISCN against the following methods are tested: CNN\cite{o2015introduction}, GoogLeNet\cite{szegedy2015going}, self-regulated network (RegNet)\cite{xu2022regnet}, Multi-Scale Binary Pattern Encoding Network (MSBPNet)\cite{vuong2021multi} and MixPro\cite{zhao2023mixpro}. It is worth noting that the performances based on MLP and SEA are compared to further verify the significance of SEA module.

\begin{itemize}
    \item CNN\cite{o2015introduction}. CNN is a traditional image processing architecture and serves as a fundamental component of multiple image classification networks. It is designed to solve image pattern recognition tasks with a simpler structure than artificial neural networks (ANNs).
    \item GoogLeNet\cite{szegedy2015going}. GoogLeNet is a deep convolutional neural network that utilizes an efficient deep neural network architecture for computer vision, called Inception, to ensure agreement with limited computational resources.
    \item RegNet\cite{xu2022regnet}. RegNet is a variant of ResNet that introduces a memory mechanism into traditional ResNet to extract potentially complementary features. This approach has shown to be effective at extracting spatiotemporal information from images.
    \item MSBPNet\cite{vuong2021multi}. MSBPNet can effectively identify and leverage the patterns of multiple scales in a cooperative and discriminative fashion within a deep neural network, providing superior capability for pathology image analysis, such as cancer classification.
    \item MixPro\cite{zhao2023mixpro}. MixPro combines MaskMix and Progressive Attention Labeling with transformer-based network as data augmentation methods, to achieve better performance and robustness.
\end{itemize}

\subsubsection{Implementation details}
The implementation details of all experiments are as follows: all the networks are trained using SGD optimizer (momentum=0.9). The batch-size is set to 16. Learning rate is set to $5\times10^{-3}$. All layers are initialized using the initialization scheme proposed by He et. al \cite{he2015delving}. A cross entropy is adopted as a loss function and we also used $\mathscr{L}_2$ regularization.The central region of 224$\times$224 pixels is cropped for all heatmaps. 
All experiments are implemented on PyTorch platform and executed on a workstation with two 4080Ti GPUs. 
All the networks are trained using an early stopping strategy with “UP criterion” \cite{prechelt1998automatic} to prevent the overfitting problem.


\subsection{Experiment results}

\subsubsection{Visualization of depth/normals integration}
\begin{figure}
\centerline{\includegraphics[width=1.0\columnwidth]{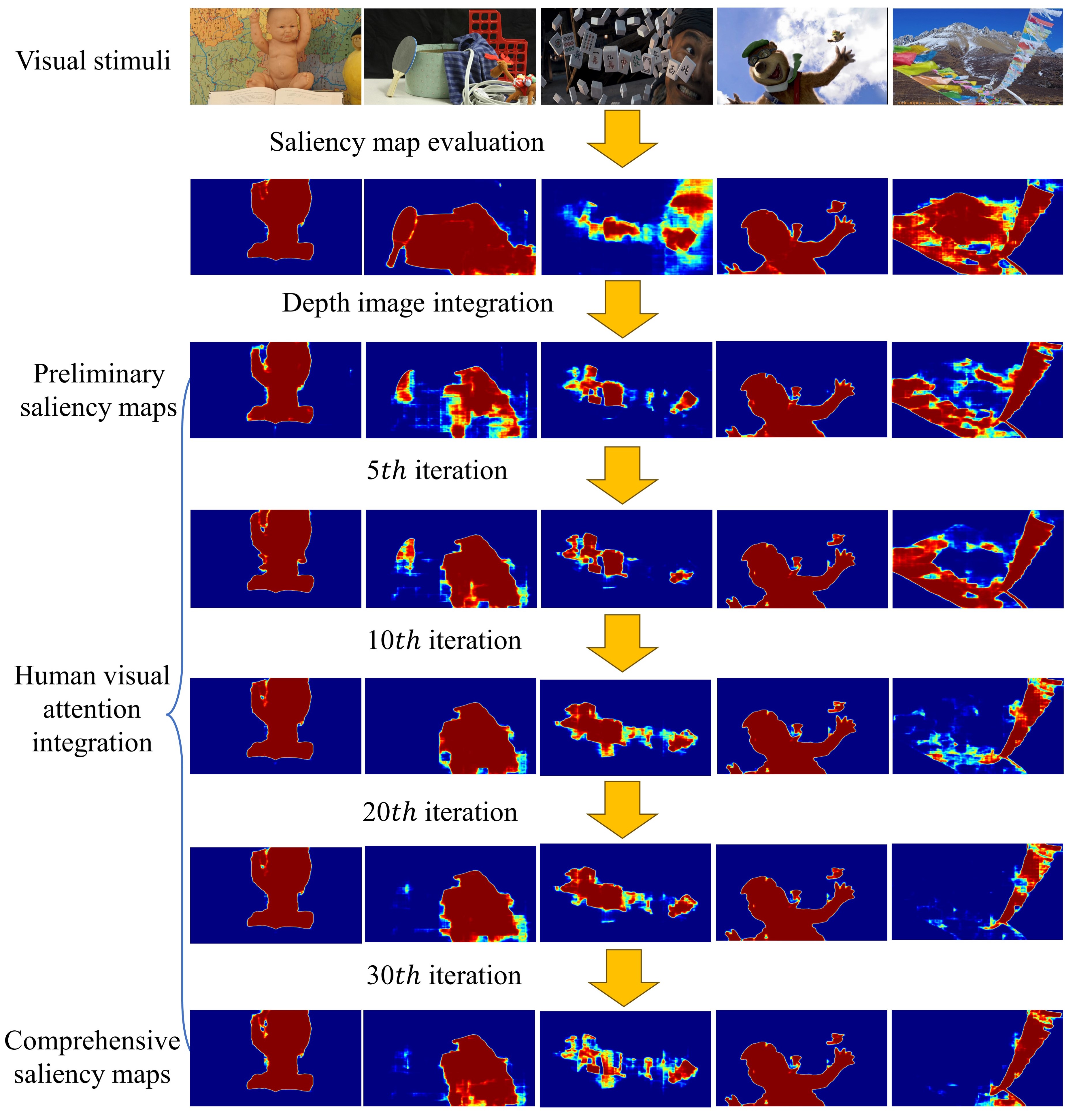}}
\caption{The processes of comprehensive visual saliency evaluation, including depth integration and normal controls integration.}
\label{iteration}
\vspace{0em}
\end{figure}

The process of comprehensive saliency maps evaluation has been introduced in Section III-C. By integrating with depth maps the evaluation of saliency maps are corrected with more details. By introducing depth maps, the saliency of objects in the background areas decreases, e.g. the human fave and mountain in the background. Moreover, by introducing visual attentions of normal controls, normal human visual attention further corrects the saliency maps and adds more details matching common visual habits of normal controls. For example, although the cartoon bear behind is judged as less salient object with the introduce of the depth map, it is seen more salient as introducing visual attention of more normal controls. The whole process is presented in Fig. \ref{iteration}.


\subsubsection{Ablation experiments}

\begin{figure}
\centerline{\includegraphics[width=1.0\columnwidth]{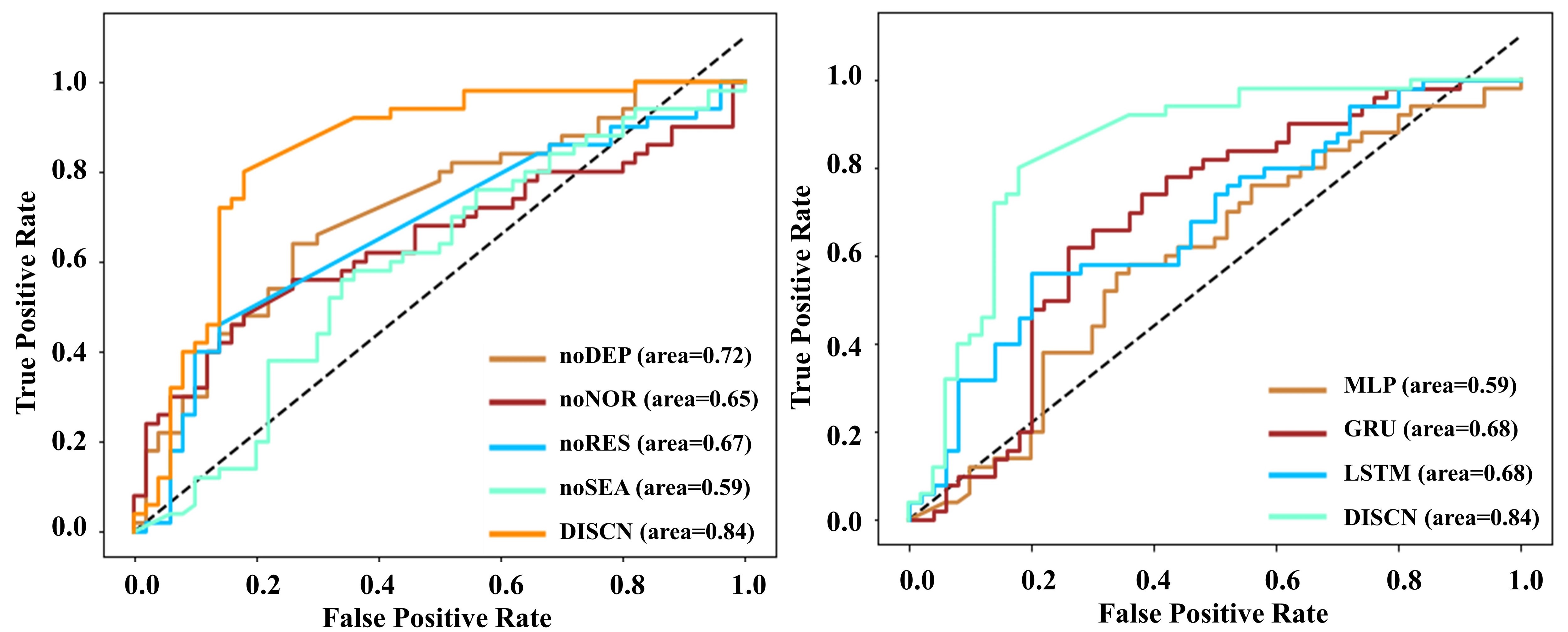}}
\caption{The ROC curves of ablation models of four modules in DISCN and four different feature fusing methods.}
\label{ablation1-f}
\vspace{0em}
\end{figure}

The results of ablation experiments are shown in Fig. \ref{ablation1-f} and TABLE \ref{ablation1-t} and \ref{ablation2-t} separately. From TABLE \ref{ablation1-t} we can see that, depth integration and normals integration contributes 0.13 and 0.17 accuracy to DISCN. The residual feature extractor contributes 0.17 accuracy, as features extracted by deep convolutional layers can not represent visual saliency adequately without features of shallow layers. SEA is the most important module as it contributes 0.23 accuracy to DISCN, which indicates that SEA is efficient for salient feature fusion.

Additionally, TABLE \ref{ablation2-t} validates performances of DISCN with four feature fusing modules. From TABLE \ref{ablation2-t} it can be concluded that recurrent structures with temporal order perform better than MLP. SEA, as a fusing structure that contains both temporal order and attention, performs the best. The ROC curves of two sets of models are plotted in Fig. \ref{ablation1-f}, which is consistent with tables and shows the superior of DISCN from statistical perspective.

\begin{table}[htbp]
    \vspace{-1em}
	\centering
	\caption{Performances of networks for evaluating effectiveness of different modules in DISCN.}
        \resizebox{\linewidth}{!}{
	\begin{tabular}{cccccc}
		\toprule  
		Metrics & noDEP & noNOR & noRES & noSEA & \textbf{DISCN} \\ 
		\midrule  
        Accuracy &         
        0.68$\pm$0.01 & 0.64$\pm$0.02 & 0.64$\pm$0.01 & 0.58$\pm$0.01&\textbf{0.81}$\pm$\textbf{0.01} \\
        
        Recall & 
        0.66$\pm$0.04 & 0.48$\pm$0.08 & 0.42$\pm$0.05 & 0.38$\pm$0.11&\textbf{0.80}$\pm$\textbf{0.01} \\
        
        Precision & 
        0.76$\pm$0.02 & 0.60$\pm$0.11 & 0.85$\pm$0.02 &0.65$\pm$0.14 &\textbf{0.86}$\pm$\textbf{0.02} \\
        
        F1-score & 
        0.66$\pm$0.01 & 0.52$\pm$0.08 & 0.50$\pm$0.04 & 0.40$\pm$0.05 &\textbf{0.81}$\pm$\textbf{0.01}  \\

        AUC & 
        0.68$\pm$0.01 & 0.64$\pm$0.02 & 0.64$\pm$0.01 & 0.58$\pm$0.01&\textbf{0.81}$\pm$\textbf{0.01}  \\
		\bottomrule  
	\end{tabular}}
 \label{ablation1-t}
\end{table}

\begin{table}[htbp]
    \vspace{-1em}
	\centering
    \small
	\caption{Performances of DISCN with four different feature fusing modules.}
        \resizebox{0.9\linewidth}{!}{
	   \begin{tabular}{ccccc}
		\toprule  
		Metrics & MLP & LSTM & GRU & SEA \\ 
		\midrule  
            
	    Accuracy &         
        0.58$\pm$0.01 & 0.65$\pm$0.01 & 0.66$\pm$0.02 & \textbf{0.81}$\pm$\textbf{0.01} \\
        
        Recall & 
        0.38$\pm$0.11 & 0.56$\pm$0.03 & 0.84$\pm$0.03 & \textbf{0.80}$\pm$\textbf{0.01} \\
        
        Precision & 
        0.65$\pm$0.14 & 0.69$\pm$0.02 & 0.64$\pm$0.02 & \textbf{0.86}$\pm$\textbf{0.02} \\
        
        F1-score & 
        0.40$\pm$0.05 & 0.60$\pm$0.02 & 0.71$\pm$0.01 & \textbf{0.81}$\pm$\textbf{0.01}  \\

        AUC & 
        0.58$\pm$0.01 & 0.65$\pm$0.01 & 0.66$\pm$0.02 & \textbf{0.81}$\pm$\textbf{0.01}  \\
		\bottomrule  
	\end{tabular}}
 \label{ablation2-t}
\end{table}

\subsubsection{Feature extracting structures}

\begin{figure}
\centerline{\includegraphics[width=1.0\columnwidth]{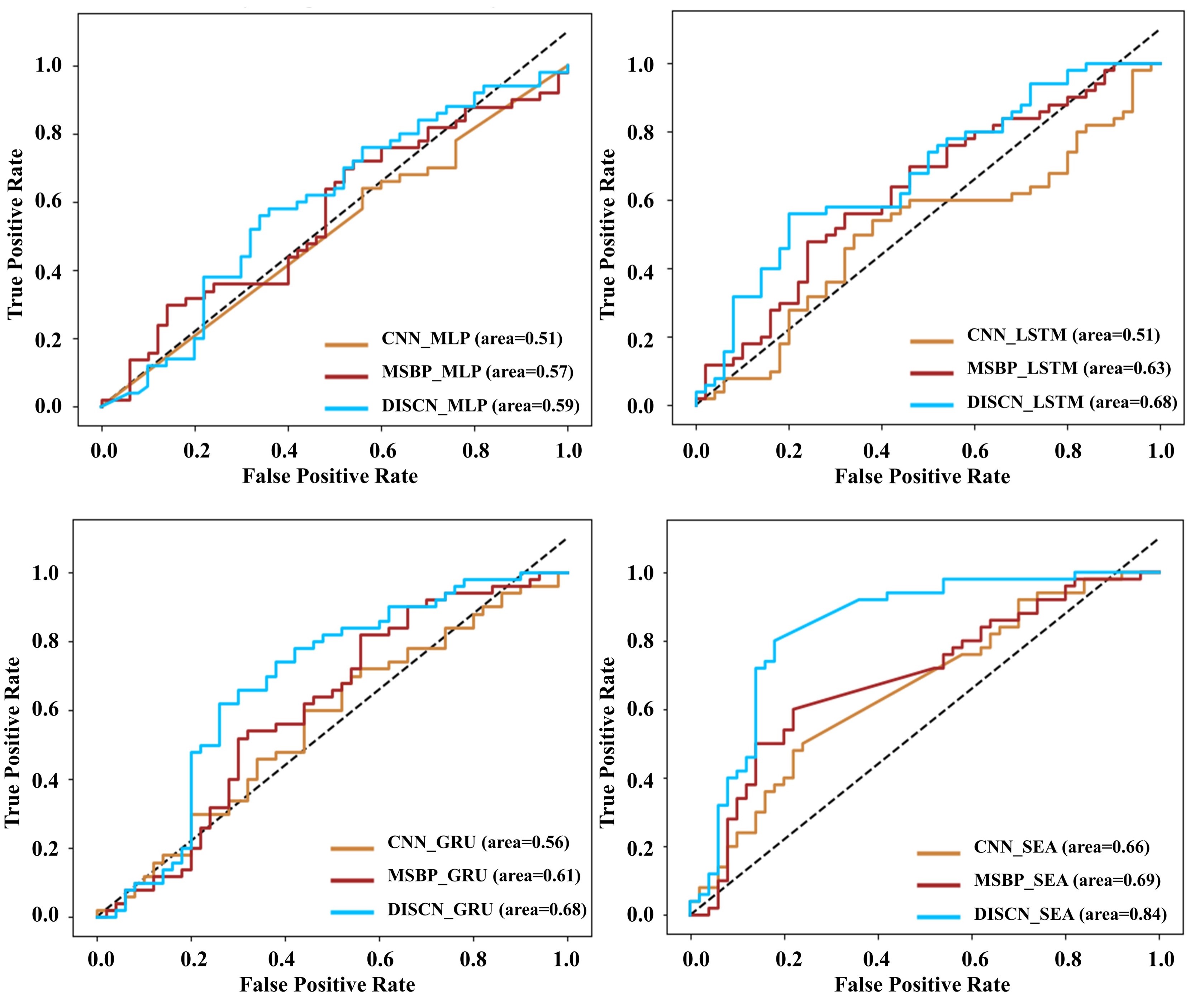}}
\caption{The ROC curves of different feature extracting structures with four different feature fusing modules.}
\label{compare1-f}
\vspace{0em}
\end{figure}

The effectiveness of three feature extracting structures with four different feature fusing modules are verified in TABLE \ref{compare1-t} and Fig. \ref{compare1-f}. In TABLE \ref{compare1-t}, three feature extracting structures perform worst with MLP, and the difference of performances are not significant. The performances of LSTM and GRU is intermediate, and the differences of performances appear more significantly. MABPNet with multi-scaled features performs better than CNN with pure convolution, and DISCN with multi-layered residual layers performs the best. As for structures with SEA, differences between models are most significant, DISCN performs the best, MSPBNet performs intermediately and CNN performs the worst. For the same feature extracting structure, it performs the best based on SEA, intermediately based on GRU and LSTM, and the worst based on MLP. Fig. \ref{compare1-f} shows consistent results with TABLE \ref{compare1-t} from statistical perspective.

\begin{table}[htbp]
    \vspace{-1em}
	\centering
    \small
	\caption{Performances of different feature extracting structures with four different feature fusing modules.}
    \scalebox{0.78}{
	\begin{tabular}{cccccc}
		\toprule  
		Feature extractor& Metrics & MLP & LSTM & GRU & SEA \\ 
		\midrule  
        
	    \multirow{5}{*}{CNN}&Accuracy &     
        0.53$\pm$0.01 & 0.54$\pm$0.01 & 0.57$\pm$0.01 & \textbf{0.63}$\pm$\textbf{0.01} \\
        &Recall & 
        0.62$\pm$0.19 & 0.32$\pm$0.14 & 0.68$\pm$0.13 & \textbf{0.50}$\pm$\textbf{0.06} \\
        &Precision & 
        0.51$\pm$0.10 & 0.43$\pm$0.15 & 0.61$\pm$0.01 & \textbf{0.73}$\pm$\textbf{0.02} \\
        &F1-score & 
        0.47$\pm$0.07 & 0.30$\pm$0.07 & 0.57$\pm$0.01 & \textbf{0.53}$\pm$\textbf{0.04}  \\
        &AUC & 
        0.53$\pm$0.01 & 0.54$\pm$0.01 & 0.56$\pm$0.03 & \textbf{0.63}$\pm$\textbf{0.01}  \\
        \midrule  

        \multirow{5}{*}{MSBPNet}&Accuracy & 
        0.59$\pm$0.01 & 0.60$\pm$0.01 & 0.66$\pm$0.01 & \textbf{0.68}$\pm$\textbf{0.01} \\
        &Recall & 
        0.72$\pm$0.09 & 0.44$\pm$0.08 & 0.70$\pm$0.04 & \textbf{0.60}$\pm$\textbf{0.02} \\
        &Precision & 
        0.61$\pm$0.01 & 0.78$\pm$0.04 & 0.70$\pm$0.01 & \textbf{0.77}$\pm$\textbf{0.02} \\
        &F1-score & 
        0.60$\pm$0.02 & 0.49$\pm$0.01 & 0.67$\pm$0.01 & \textbf{0.64}$\pm$\textbf{0.01}  \\
        &AUC & 
        0.59$\pm$0.01 & 0.60$\pm$0.01 & 0.66$\pm$0.01 & \textbf{0.68}$\pm$\textbf{0.01}  \\

        
        \midrule  

        \multirow{5}{*}{DISCN}&Accuracy &
        0.58$\pm$0.01 & 0.79$\pm$0.06 & 0.66$\pm$0.02 & \textbf{0.81}$\pm$\textbf{0.01} \\
        &Recall & 
        0.38$\pm$0.11 & 0.71$\pm$0.01 & 0.84$\pm$0.03 & \textbf{0.80}$\pm$\textbf{0.01} \\
        &Precision & 
        0.65$\pm$0.14 & 0.86$\pm$0.02 & 0.64$\pm$0.02 & \textbf{0.86}$\pm$\textbf{0.02} \\
        &F1-score & 
        0.40$\pm$0.05 & 0.77$\pm$0.01 & 0.71$\pm$0.01 & \textbf{0.81}$\pm$\textbf{0.01}  \\
        &AUC & 
        0.58$\pm$0.01 & 0.79$\pm$0.07 & 0.66$\pm$0.02 & \textbf{0.81}$\pm$\textbf{0.01}  \\
		\bottomrule  
	\end{tabular}}
 \label{compare1-t}
\end{table}

\subsubsection{Comparison experiments}

\begin{figure}
\centerline{\includegraphics[width=1.0\columnwidth]{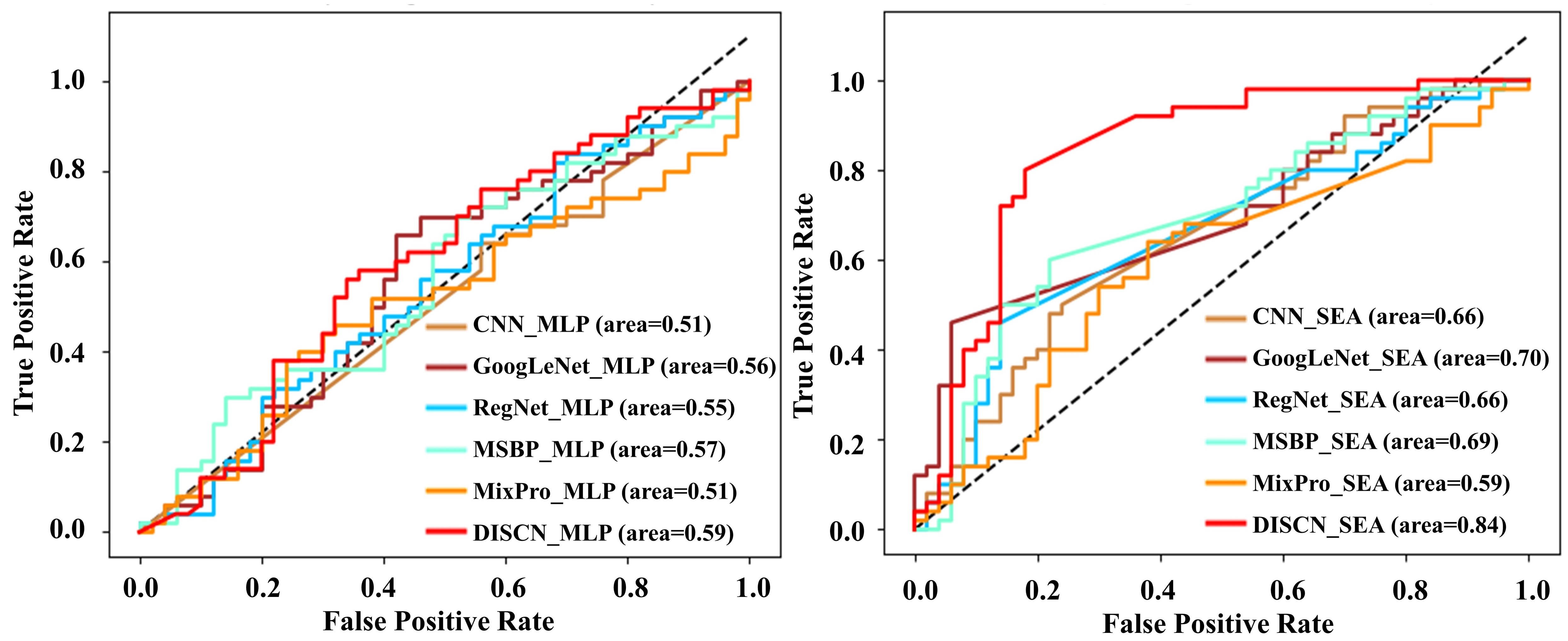}}
\caption{The ROC curves of DISCN and comparison models with MLP and SEA feature fusing modules.}
\label{compare2-f}
\vspace{0em}
\end{figure}

TABLE \ref{compare2-t} and Fig. \ref{compare2-f} depict the detailed performances of all state-of-the-art methods and the proposed DISCN. The following observations can be made from TABLE \ref{compare2-t} and Fig. \ref{compare2-f}: 1) Our proposed DISCN outperforms all state-of-the-art methods equipped with two different feature fusing modules by a significant margin. 2) CNN and GoogLeNet, as convolutional feature extracting structures, use relatively shallower features for classification. The two models perform better based on SEA than MLP, which proves that it is feasible to use simple convolutional features of image for classification, and the temporal attention mechanism in SEA can better improve network performance; 3) RegNet and MixPro are heavyweight networks and use complex image features. They perform the worst, indicating that overly complex networks may not be suitable for simple eye movement feature extraction due to issues like overfitting. 4) MSBPNet is also heavyweighted but outperforms RegNet and MixPro, which is benifited by the residual like multi-scaled feature integration mechanism; 5) Models that use SEA module consistently perform better than MLP, as SEA introduces temporal orders of salient features, and focuses on information from feature vectors that is critical for AD diagnosis, enabling the model to make more accurate judgments.

\begin{table}[htbp]
	\centering
    \small
	\caption{Performances of combinations with different feature extracting modules and integrating modules.}
	\scalebox{0.9}{\begin{tabular}{ccccc}
		\toprule  
		& Metrics & MLP & SEA & Improvement\\ 
		\midrule  
         \multirow{5}{*}{CNN\cite{2010CNN}}
		& Accuracy & 0.53$\pm$0.01 & 0.63$\pm$0.01 & 0.10$\uparrow$\\
        & Recall & 0.62$\pm$0.19 &  0.50$\pm$0.06 &0.12$\downarrow$ \\
        & Precision & 0.51$\pm$0.10 & 0.73$\pm$0.02 &0.22$\uparrow$ \\
        & AUC & 0.53$\pm$0.01 &0.63$\pm$0.01 & 0.10$\uparrow$\\
        & F1-score & 0.47$\pm$0.07 & 0.53$\pm$0.04 &0.06$\uparrow$ \\
        
        \midrule  
        \multirow{5}{*}{GoogLeNet\cite{szegedy2015going}}
		& Accuracy & 0.60$\pm$0.01 & 0.69$\pm$0.01 & 0.09$\uparrow$\\
        & Recall & 0.66$\pm$0.08 & 0.46$\pm$0.06 &0.20$\downarrow$\\
        & Precision & 0.69$\pm$0.04 & 0.90$\pm$0.02 &0.21$\uparrow$\\
        & AUC & 0.60$\pm$0.01 &0.69$\pm$0.01 & 0.09$\uparrow$\\
        & F1-score & 0.61$\pm$0.01 & 0.56$\pm$0.03 &0.05$\downarrow$\\
        
        \midrule  
        \multirow{5}{*}{RegNet\cite{xu2022regnet}}
		& Accuracy & 0.56$\pm$0.01 & 0.66$\pm$0.01 &0.10$\uparrow$\\
        & Recall & 0.82$\pm$0.07 & 0.46$\pm$0.05 &0.36$\downarrow$\\
        & Precision & 0.55$\pm$0.01 & 0.81$\pm$0.01 &0.26$\uparrow$\\
        & AUC & 0.56$\pm$0.01 &0.66$\pm$0.01 & 0.10$\uparrow$\\
        & F1-score & 0.63$\pm$0.02 & 0.55$\pm$0.02 &0.10$\downarrow$\\

        \midrule  
        \multirow{5}{*}{MSBPNet\cite{vuong2021multi}}
		& Accuracy & 0.59$\pm$0.01 & 0.68$\pm$0.01 &0.09$\uparrow$\\
        & Recall & 0.72$\pm$0.09 & 0.60$\pm$0.02 &0.12$\downarrow$\\
        & Precision & 0.61$\pm$0.01 & 0.77$\pm$0.02 &0.16$\uparrow$\\
        & AUC & 0.59$\pm$0.01 & 0.68$\pm$0.01 &0.07$\uparrow$ \\
        & F1-score & 0.60$\pm$0.02 & 0.64$\pm$0.01 &0.04$\uparrow$\\
        
        \midrule  
        \multirow{5}{*}{MixPro\cite{zhao2023mixpro}}
		& Accuracy & 0.53$\pm$0.01 &0.61$\pm$0.01 &0.08$\uparrow$\\
        & Recall & 0.30$\pm$0.14 & 0.68$\pm$0.07 &0.38$\uparrow$\\
        & Precision & 0.43$\pm$0.15 & 0.69$\pm$0.04 &0.27$\uparrow$\\
        & AUC & 0.53$\pm$0.01 & 0.61$\pm$0.01 & 0.08$\uparrow$\\
        & F1-score & 0.27$\pm$0.07 & 0.61$\pm$0.02 &0.34$\uparrow$\\

        

        \midrule  
        \multirow{5}{*}{\textbf{DISCN}}
		& Accuracy & 0.58$\pm$0.01 &\textbf{0.81}$\pm$\textbf{0.01} &0.23$\uparrow$\\
        & Recall & 0.38$\pm$0.11 & \textbf{0.80}$\pm$\textbf{0.01} & 0.42$\uparrow$\\
        & Precision & 0.65$\pm$0.14 &\textbf{0.86}$\pm$\textbf{0.02} &0.21$\uparrow$\\
        & AUC & 0.58$\pm$0.01& \textbf{0.81}$\pm$\textbf{0.01} & 0.23$\uparrow$\\
        & F1-score & 0.40$\pm$0.05 &\textbf{0.81}$\pm$\textbf{0.01} &0.41$\uparrow$\\
		\bottomrule  
	\end{tabular}}
 \label{compare2-t}
\end{table}

\section{Conclusion}

In this paper, we propose a novel approach for deep learning-based Alzheimer's disease (AD) diagnosis called DISCN, which jointly utilizes eye movement and visual stimuli. By leveraging the advantages of multi-image integration module and serial attention module, we effectively integrate visual stimuli and heatmaps that contain eye movements for a more comprehensive visual saliency evaluation. We extract features related to the cognitive impairment of AD patients and send them to the serial attention module for diagnosis. Our experimental results demonstrate that coherent visual saliency of visual stimuli and eye movements is proposing to be used for AD diagnosis. Our network consistently outperforms state-of-the-art methods based on the visual saliency dataset.


\section*{Acknowledgement}
The authors would like to thank all of the study subjects and their families that participated in the data collection. Sincerely thanks to the reviewers for their efforts in reviewing and improving my paper.
\bibliographystyle{IEEEtran}
\bibliography{reference}

\end{document}